\documentclass{IEEEtaes}

\usepackage{color,array,amsthm}
\usepackage{graphicx}
\usepackage{amsfonts}
\usepackage{amsmath}
\usepackage{algorithm}
\usepackage{algorithmic}

\usepackage{float}
\usepackage{subfigure}

\jvol{XX}
\jnum{XX}
\jmonth{XXXXX}
\paper{1234567}
\pubyear{2022}
\doiinfo{TAES.2022.Doi Number}

\newcommand{\tabincell}[2]{\begin{tabular}{@{}#1@{}}#2\end{tabular}}

\begin{document}

\title{Learning with Geometry: Including Riemannian Geometric Features in Coefficient of Pressure Prediction on Aircraft Wings}

\author{Liwei Hu\href{https://orcid.org/0000-0003-4994-9252}{\includegraphics[scale=0.5]{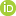}}}
\member{Student member, IEEE}
\affil{Hebei University of Science and Technology, Shijiazhaung, China}

\author{Wenyong Wang\href{https://orcid.org/0000-0003-4095-547X}{\includegraphics[scale=0.5]{ORCIDiD.png}}}
\member{Senior member, IEEE}
\affil{University of Electronic Science and Technology of China, Chengdu, China}

\author{Yu Xiang\href{https://orcid.org/0000-0001-9622-7661}{\includegraphics[scale=0.5]{ORCIDiD.png}}}
\member{Member, IEEE}
\affil{University of Electronic Science and Technology of China, Chengdu, China}

\author{Stefan Sommer\href{https://orcid.org/0000-0001-6784-0328}{\includegraphics[scale=0.5]{ORCIDiD.png}}}
\affil{University of Copenhagen, Copenhagen, Denmark} 


\receiveddate{Manuscript received XXXXX 00, 0000; revised XXXXX 00, 0000; accepted XXXXX 00, 0000.\\
This paper is a further study of DOI:10.1109/TAES.2023.3276735.\\
This work is supported by the National Natural Science Foundation of China No. 62250067.\\}

\corresp{Corresponding author: Stefan Sommer and Yu Xiang}

\authoraddress{
L. Hu, lecturer, is with the School of Information Science and Engineering, Hebei University of Science and Technology, Shijiazhuang 050091, China. (e-mail: \href{liweihu@hebust.edu.cn}{liweihu@hebust.edu.cn}).\\ 
L. Hu, was with the School of Computer Science and Engineering, University of Electronic Science and Technology of China, Chengdu 611731, China. (e-mail: \href{liweihu@std.uestc.edu.cn}{liweihu@std.uestc.edu.cn}).\\
L. Hu, visitor, was with the Department of Computer Science, University of Copenhagen.\\
W. Wang, professor, is with the School of Computer Science and Engineering, University of Electronic Science and Technology of China, Chengdu 611731, China. (e-mail: \href{wangwy@uestc.edu.cn}{wangwy@uestc.edu.cn}).\\
W. Wang, special-term professor, is with the International Institute of Next Generation Internet, Macau University of Science and Technology, Macau 519020, China.\\ 
Y. Xiang, associate professor, is with the School of Computer Science and Engineering, University of Electronic Science and Technology of China, Chengdu 611731, China. (e-mail: \href{jcxiang@uestc.edu.cn}{jcxiang@uestc.edu.cn}).\\
S. Sommer, professor, is with the Department of Computer Science, University of Copenhagen. (e-mail: \href{sommer@di.ku.dk}{sommer@di.ku.dk}).}

\editor{}
\supplementary{}

\markboth{Hu ET AL.}{Including Riemannian Geometric Features in Coefficient of Pressure Prediction on Aircraft Wings}
\maketitle

\begin{abstract}
We propose to incorporate Riemannian geometric features from the geometry of aircraft wing surfaces in the prediction of coefficient of pressure ($C_P$) on the aircraft wing. Contrary to existing approaches that treat the wing surface as a flat object, we represent the wing as a piecewise smooth manifold and calculate a set of Riemannian geometric features (Riemannian metric, connection, and curvature) over points of the wing. Combining these features in neighborhoods of points on the wing with coordinates and flight conditions gives inputs to a deep learning model that predicts $C_{P}$ distributions. Experimental results show that the method with incorporation of Riemannian geometric features, compared to state-of-the-art Deep Attention Network (DAN), reduces the predicted mean square error (MSE) of $C_{P}$ by an average of 15.00\% for the DLR-F11 aircraft test set.
\end{abstract}


\section{INTRODUCTION}
E{\scshape xisting} deep learning methods that treat 3D objects in the real world as flat Euclidean data utilize 2D images or slices of a 3D object as inputs to learn neural representations to reflect its geometry \cite{yang2023inverse,zahn2023prediction}. These 3D objects, such as 3D aircraft shape \cite{zuo2023fast}, face structure \cite{zhang2024leave} and point cloud \cite{yu2022point}, etc. have intrinsic geometric properties that can be measured within the object itself without any reference to the space it may be located in \cite{baker2024learning}. However, the discretization of 2D images and slices may lose intrinsic geometric properties of 3D objects \cite{alwadee2024latup,liu2024simultaneous}. Consequently, the neural representations learned from 2D images or slices of a 3D object can not represent its intrinsic geometric integrity accurately.

Recent studies have shown that non-Euclidean space can be used to accurately represent objects with intricate geometries \cite{zhang2024data}. For examples, hyperbolic space \cite{mettes2024hyperbolic} and spheres \cite{cao2024knowledge}, two type of Riemannian manifolds, were used to represent data with tree-like geometry and cyclical geometry, respectively. In aerodynamics, there is a growing trend in treating aircraft geometries as non-Euclidean data. Deng and Wang et al. show that airfoils marked by 2D coordinate points can be represented as 1D curves (non-Euclidean manifolds) embedded in $\mathbb R^2$ \cite{deng2023prediction, wang2023airfoil}. Further more, Xiang et al. demonstrated for the first time that the Riemannian metric, calculated from airfoil manifolds, can be used to help deep learning models improve the prediction accuracy of aerodynamic coefficients \cite{xiang2023manifold}. However, the above studies only focus on airfoils represented by 1D curves in $\mathbb R^2$, and the aerodynamic coefficient prediction for an arbitrary point is based on the Riemannian geometric feature calculated at the point, ignoring the geometric features calculated from neighbors around the point.

Motivated by manifold theory, we propose a Riemannian geometric features-incorporated learning (RGFiL) method \footnote{The source code of our work can be found at https://github.com/huliwei123/RGFiL.} and apply it to coefficient of pressure ($C_P$) predictions on the wing of a 3D aircraft (in this paper, a 3D aircraft is represented as a 2D curved surface embedded in $\mathbb R^3$). RGFiL constructs a piecewise smooth manifold for a given 3D aircraft, and then a set of Riemannian geometric features (Riemannian metric, connection, and curvature) at an arbitrary point as well as its 8 neighbors are all included into consideration to represent the geometry of the aircraft. In this paper, the geometric features calculated from Riemannian manifolds are called Riemannian geometric features. At last, Riemannian geometric features, flight conditions and coordinates are input to a neural network to learn neural representations for predicting $C_{P}$ on the wing of the aircraft.

We highlight the contributions of this paper as follows: 1) This paper proposes a novel deep learning model that incorporates Riemannian geometric features for predicting $C_{P}$ distributions; 2) Experimental results show that our model has better performances in predicting $C_{P}$ distributions than existing methods without geometric features, meaning that the Riemannian geometric features are well recognized and utilized by the method; 3) The method predicts $C_{P}$ value according to the geometry of small neighborhoods on the curved surface of a 3D aircraft rather than the geometry at a single point.

The remainder of this paper is organized as follows. Section \ref{section_related_work} provides an overview of learning methods for aerodynamic coefficient predictions based on two different aircraft geometry representations. Section \ref{section_methodology} introduces the details of RGFiL. Section \ref{section_experimental_result} shows the experimental details and results of five deep learning models on DLR-F11 aircraft dataset. The conclusions are presented in Section \ref{section_conclusion}.

\section{RELATED WORKS}
\label{section_related_work}
In this section, we discuss the state-of-the-arts in the field of aircraft shape representations for aerodynamic coefficient predictions. Specifically, we focus on aerodynamic coefficient predictions based on aircraft image (slice) shape representations and aerodynamic coefficient predictions based on aircraft manifold shape representations.

\subsection{Aerodynamic coefficient predictions based on aircraft image (slice) shape representations} 

This method commonly organize the shape of aircrafts into multiple images or 2D slices and input them into deep learning models to obtain massive latent neural representations for aircraft shapes \cite{catalani2023comparative,jacob2021deep,li2021deep}. Zuo \cite{zuo2023fast} et al. proposed a Deep Attention Network (DAN) model to extract neural representations from airfoil dataset provided by University of Illinois Urbana-Champaign (UIUC) \cite{selig2011uiuc}. The core concept of DAN is to generate latent neural representations from airfoil images by utilizing a Vision Transformer (VIT) \cite{han2022survey}, and then taking the neural representations and flight conditions of airfoils as inputs of a Multi-Layer Perceptron (MLP) submodule to predict the $C_P$ distribution on airfoil curves. Li et al. used a Deep Convolution neural network (DCNN)-based geometric filtering model to identify abnormalities in airfoils generated by Wasserstein GAN (WGAN), and further to predicted the $C_P$ distributions for normal airfoils. Notably, all input airfoils were structured as images, and the abnormalities detection and the $C_P$ prediction are achieved through the classification and regression of DCNN \cite{li2021deep}. Qu et al. divided a wing in 3D space into multiple 2D slices and used a DCNN to predict $C_{P}$ distribution for each slice \cite{qu2022predicting}. Similar studies can be found in \cite{lei2021deeplearning,xiong2023point,zhang2023airfoil}.

The above studies predict $C_{P}$ distributions for airfoils or 2D slices of 3D aircrafts. However, the discretization of airfoils or 2D slices may lose intrinsic geometric properties of 3D aircrafts.

\subsection{Aerodynamic coefficient predictions based on aircraft manifold shape representations}
In this method, the curved geometry of a 3D aircraft is typically mapped into a low dimensional space \cite{chen2022learning,van2022effect}, and the geometry representations of the aircraft are expressed by calculating Riemannian geometric features based on parameters defined in the low dimensional space. 

Wang et al. used cubic splines to establish a mapping between 2D Cartesian coordinates of UIUC airfoils and 1D curvilinear coordinates \cite{wang2021airfoil}. Deng et al. utilized a univalent transformation to convert the geometrical information of airfoil mesh from 2D Cartesian coordinates to curvilinear coordinates \cite{deng2023prediction}. Although, the curvilinear coordinates are built based on manifold theory, there is no fundamental difference between using curvilinear coordinates and using Cartesian coordinates to represent geometries, because there is no further feature extractions. Xiang et al. constructed a 1D manifold for a given 2D UIUC airfoil, and further took the calculated Riemannian metric, a Riemannian geometric feature, as input to a multi-feature learning method to predict coefficient of Drag ($C_{D}$), which results in a significant reduction compared to using images (slices) as inputs \cite{xiang2023manifold}.  

The above studies were carried out on 2D airfoils, and to the best of our knowledge, there has been no studies on constructing manifolds for wing surfaces in 3D space and extracting further geometric features for them.

In summary, compared with aerodynamic coefficient predictions based on aircraft image (slice) shape representations, aerodynamic coefficient predictions based on aircraft manifold shape representations can further reduce the prediction errors of aerodynamic coefficients. In addition, Riemannian geometric features calculated from aircraft manifolds can preserve the intrinsic geometric properties of aircraft shapes theoretically. However, existing studies focused on shape representations for 2D airfoils, and the challenge of this method is how to extract geometric features for wings in 3D space.

\section{Method}
\label{section_methodology}

\begin{figure*}[t]
\centering
\includegraphics[scale=0.6]{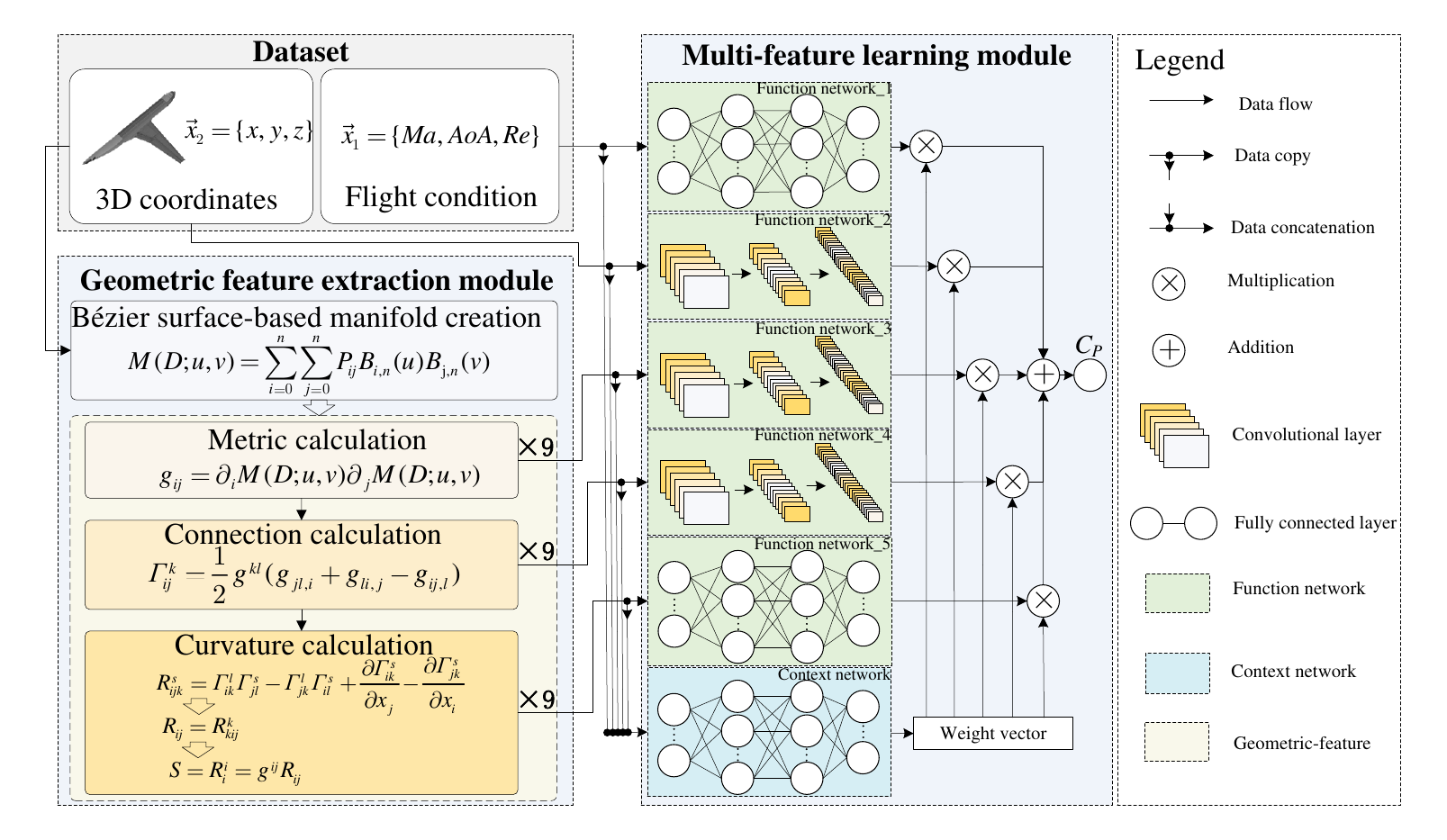}
\caption{The structure of RGFiL.}
\label{fig_structure}
\end{figure*}

In Figure. \ref{fig_structure}, RGFiL consists of two submodules: a geometric feature extraction module and a multi-feature learning module. In the geometric feature extraction module, a set of self-intersection-free Bézier surfaces is used to approximate a 3D aircraft to form a piecewise smooth manifold, and then Riemannian metric $g_{ij}$, connection $\mit\Gamma_{i j}^k$ and curvature $S$ are calculated as a set of Riemannian geometric features to represent the geometry of the aircraft. Because the $C_{P}$ value at an arbitrary point is related to a neighborhood around the point, the Riemannian geometric features of the point as well as its 8 neighbors are all considered. In the multi-feature learning module, the flight condition $\vec{\mathbf{x}}_1$, the coordinate points $\vec{\mathbf{x}}_2$ and Riemannian geometric features related to 9 points are taken as inputs to predict $C_{P}$ on wing of the 3D aircraft. The multi-feature learning structure involves discovering importance of neural representations, which can be used to verify the effectiveness of Riemannian geometric features \cite{xiang2023manifold,hu2022aerodynamic}.

\subsection{The Geometric Feature Extraction Module}

\subsubsection{Manifold Construction}
We use a set of self-intersection-free \cite{xiang2023manifold} Bézier surfaces that approximates a 3D aircraft to form a piecewise smooth manifold (we call each part of the piecewise smooth manifold a segment). 

To simplify the description, we focus on a specific segment. Consider a set of control points:
\begin{equation}
\nonumber
\label{equ_dataset_matrix}
D=
\begin{bmatrix}
  P_{00} & P_{01} &\cdots & P_{0n} \\
  P_{10} & P_{ab} &\cdots & P_{1n} \\
  \vdots & \vdots &\ddots & \vdots \\  
  P_{m0} & P_{m1} &\cdots & P_{mn} 
\end{bmatrix}
\end{equation} 
where $P_{ab}=(x_{ab},y_{ab},z_{ab})$, $a=1,2,\cdots,m$ and $b=1,2,\cdots,n$. A Bézier surface can be built:

\begin{equation}
\label{equ_bezier}
\begin{cases}
  F(u, v)=\sum_{a=0}^m \sum_{b=0}^n P_{a b} B_{a, m}(u) B_{b, n}(v) \\
  B_{a, m}(u)=\frac{m !}{a !(m-a) !} u^{a}(1-u)^{m-a}\\
  B_{b, n}(v)=\frac{n !}{b !(n-b) !} v^{b}(1-v)^{n-b}
\end{cases}
\end{equation}
where $F(u,v)$ is a Bézier surface function, $u$ and $v$ $(u,v \in [0,1])$ are two parameters of $F(u,v)$, $m$ and $n$ are the degrees of $F(u,v)$, and $B_{a, m}(u)$ and $B_{b,n}(v)$ are coefficients of $F(u,v)$.

The Jacobian matrix of $F(u,v)$ is:
\begin{equation}
\label{equ_jacobin_F}
J_F=d(F(u,v))=
\begin{bmatrix}
  \frac{\partial F_x}{\partial u} & \frac{\partial F_x}{\partial v} \\[6pt]
  \frac{\partial F_y}{\partial u} & \frac{\partial F_y}{\partial u} \\[6pt]
  \frac{\partial F_z}{\partial u} & \frac{\partial F_z}{\partial u}
\end{bmatrix}.
\end{equation}

In this case, the Rank of $J_F$ is 2. $F(u,v)$ is an injection \cite{lee2012smooth} (i.e., $F(u,v)$ is a smooth immersion \cite{RiemannianChen}). If the function $F(u,v)$ has no self-intersections, then $F(u,v)$ is a smooth embedding in $\mathbb{R}^3$ \cite{lee2012smooth}. Multiple smooth embeddings are concatenated together to form a piecewise smooth manifold $\mathcal{M}$.

\subsubsection{Calculation of Riemannian Geometric Features}
1) Riemannian metric: A Riemannian metric $g_{ij}$ defined at an arbitrary point $P_{ab} \in \mathcal{M}$, is a smooth covariant 2-tensor field representing an inner product on the tangent space of $\mathcal{M}$. Let $x_1=u, x_2=v$, the Riemannian metric can be calculated as
\begin{equation}
\label{equ_geometric_feature}
\begin{aligned}
g_{ij}=&\langle \frac{\partial}{\partial x_{i}} F(u,v), \frac{\partial}{\partial x_{j}} F(u,v) \rangle \\   
      =&\partial_{i} F(u,v) \partial_{j} F(u,v)
\end{aligned}
\end{equation}
where $i,j=\left\{1,2\right\}$. 

2) Connection: The connection $\nabla_{X}Y$ is a derivative of a vector field $Y$ on $(\mathcal{M},g)$ in the direction of another vector field $X$ \cite{van2023stochastic}. Suppose that $X = v^i(x) \partial_{x^i}$ and $Y = w^i(x) \partial_{x^i}$, then

\begin{equation}
\label{equ_connection}
\nabla_{X}Y = \left(X\left(w^k\right)+v^i w^j \Gamma^k_{i j}\right) \partial_{x^k}.
\end{equation}

The connection coefficient $\Gamma^k_{i j}$, the Christoffel symbols \cite{khatsymovsky2019discrete}, can be written as \cite{pennec2019riemannian}

\begin{equation}
\label{equ_gamma}
\begin{aligned}
\Gamma^k_{i j}&=\frac{1}{2} g^{k l}\left(\partial_{x^i} g_{j l}+\partial_{x^j} g_{l i}-\partial_{x^l} g_{i j}\right)\\
                           &=\frac{1}{2} g^{k l}\left(g_{j l, i}+g_{l i, j}-g_{i j, l}\right)\\
\end{aligned}
\end{equation}
where $g^{k l}$ denotes the inverse matrix of $g_{k l}$, and $i,j,k,l=\left\{1,2 \right\}$.

3) Curvature: Curvature is a measure of whether a Riemannian manifold $(\mathcal{M},g)$ is flat or curved. Curvature is defined in the form of covariant derivative by evaluation on three vector fields $X$, $Y$, $Z$:
\begin{equation}
\label{equ_curvature_definition}
  R(X, Y) Z=\nabla_X \nabla_Y Z-\nabla_Y \nabla_X Z-\nabla_{[X, Y]} Z
\end{equation}
where Lie bracket $[X, Y]$ \cite{frost2023lie} denotes the anticommutativity of $X$ and $Y$. Suppose that $X = v^i(x) \partial_{x^i}$, $Y = w^i(x) \partial_{x^i}$ and $Z = u^i(x) \partial_{x^i}$, the curvature $R(X,Y)Z$ can also be written as

\begin{equation}
\label{equ_curvature}
\begin{aligned}
  R(X, Y) Z = v^i(x) w^j(x) u^k(x) R_{ijk}^s \partial_{x^s}.
\end{aligned}
\end{equation}

Associated with (\ref{equ_curvature_definition}) and (\ref{equ_curvature}), we get the curvature coefficient $R_{ijk}^s$:
\begin{equation}
\label{equ_curvature_ceofficient}
  R_{i j k}^s=\left(\Gamma_{i k}^l \Gamma_{j l}^s-\Gamma_{j k}^l \Gamma_{i l}^s\right)+\frac{\partial \Gamma_{i k}^s}{\partial x_j}-\frac{\partial \Gamma_{j k}^s}{\partial x_i}
\end{equation}
where $i,j,k,s=\left\{1,2 \right\}$. Usually, the Ricci curvature $R_{i j}=R_{k i j}^k$ and scalar curvature $S=g^{i j} R_{i j}$ \cite{munteanu2023comparison} are used to simplify $R_{i j k}^s$.

\begin{figure}[h]
  \centering
  \includegraphics[scale=1.0]{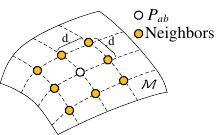}
  \caption{Neighbor selection.}
  \label{fig_selection_neighbors}
\end{figure}

4) Riemannian geometric features for 9 points: RGFiL takes Riemannian geometric features of $P_{ab}$ and its 8 neighbors as inputs to the multi-feature learning module. Figure. \ref{fig_selection_neighbors} shows how to select the 8 neighbors for $P_{ab}$. All points are on $\mathcal{M}$, and the distance $d$ between points is a hyperparameter. If the value of $d$ is relatively small, the Riemannian geometric features of all points are relatively similar.

\subsection{The Multi-feature Learning Module}

\subsubsection{Structure}
The multi-feature learning module consists of five function networks and a context network. 

The function network\_1 learns a nonlinear mapping between flight conditions $\vec{\mathbf{x}}_1=[Ma, AoA, Re]$ and $C_{P}$, where $Ma$ denotes the incoming Mach number, $AoA$ denotes the Angle of Attack, $Re$ denotes the Reynolds number. The function network\_2 learns a nonlinear mapping between coordinate $\vec{\mathbf{x}}_2=\left\{x,y,z \right\}$ on an aircraft and $C_{P}$. The function network\_3 learns a nonlinear mapping between Riemannian metric $\vec{\mathbf{x}}_{3}=\left\{g_{ij}\right\} $ and $C_{P}$. The function network\_4 learns a nonlinear mapping between connection $\vec{\mathbf{x}}_{4}=\left\{\mit\Gamma_{i j}^k\right\}$ and $C_{P}$. The function network\_5 learns a nonlinear mapping between scalar curvature $\vec{\mathbf{x}}_{5}=\left\{S\right\}$ and $C_{P}$. The context network takes $\vec{\boldsymbol{\xi}}=[\vec{\mathbf{x}}_1, \vec{\mathbf{x}}_2,\vec{\mathbf{x}}_3, \vec{\mathbf{x}}_4,\vec{\mathbf{x}}_5]$ as input to learn weight vectors for function network outputs, where $[\cdot]$ denotes the concatenation of vectors.

The output of this module is:
\begin{equation}
\label{equal_y}
 \hat{C}_{P}= \sum_{z=1}^{5} f_{z \alpha}(\vec{\mathbf{x}}_{z}) \times c_{\left(z-1\right)K+\alpha}(\vec{\boldsymbol{\xi}})
\end{equation}
where $\hat{C}_{P}$ denotes the predicted value, $f_{z\alpha}(\vec{\mathbf{x}}_{z})$ denotes the $\alpha$th component of the output vector from function network\_$z$, $K$ denotes the number of output nodes in each function networks, $c_{\alpha}(\vec{\boldsymbol{\xi}})$ denotes the $\alpha$th component of the weight vector generated by the context network.

\subsubsection{Training Method}
The loss function of this submodule is:

\begin{equation}
\label{equ_function_loss}
L=\frac{1}{M} \sum_{\beta=1}^{M}\left(C_{P_{tr}}^\beta-\hat{C}_{P}^\beta\right)^2
\end{equation}
where $M$ denotes the number of data points in training set, $C_{P_{tr}}^\beta$ denotes the real $C_P$ value of the $\beta$th sample in the training set, and $\hat{C}_{P}^\beta$ denotes the predicted value of the $\beta$th sample in the training set. The training algorithm of this module is shown as in Algorithm. \ref{alg_1}.

\begin{algorithm}
\caption{The training algorithm of the multi-feature learning module.}
\label{alg_1}
\textbf{Input}: $\vec{\mathbf{x}}_1$, $\vec{\mathbf{x}}_2$, $\vec{\mathbf{x}}_3$, $\vec{\mathbf{x}}_4$, $\vec{\mathbf{x}}_5$ and $C_{P_{tr}}$\\
\textbf{Parameter}:  learning rate $\eta$, function network parameters $\boldsymbol{\theta}_{f}$, context network parameters $\boldsymbol{\theta}_{c}$\\
\textbf{Output}: $\boldsymbol{\theta}_{f}$ and $\boldsymbol{\theta}_{c}$
\begin{algorithmic}[1] 
\STATE Initialize $\eta$, $\boldsymbol{\theta}_{f}$, $\boldsymbol{\theta}_{c}$, $e \gets 0$
\WHILE{$e <$  training iterations}
    \STATE$\vec{\boldsymbol{\xi}} \gets [\vec{\mathbf{x}}_1, \vec{\mathbf{x}}_2,\vec{\mathbf{x}}_3, \vec{\mathbf{x}}_4,\vec{\mathbf{x}}_5]$
    \STATE$\hat{C}_{P} \gets \sum_{z=1}^{5} f_{z \alpha}(\vec{\mathbf{x}}_{z}) \times c_{\left(z-1\right)K+\alpha}(\vec{\boldsymbol{\xi}})$
    \STATE$L \gets \frac{1}{M} \sum_{\beta=1}^{M}\left(C_{P_{tr}}^\beta-\hat{C}_{P}^\beta\right)^2$
    \STATE$\boldsymbol{\theta}_{f} \gets \boldsymbol{\theta}_{f}-\eta \frac{\partial L}{\partial \boldsymbol{\theta}_{f}}$
    \STATE$\boldsymbol{\theta}_{c} \gets \boldsymbol{\theta}_{c}-\eta \frac{\partial L}{\partial \boldsymbol{\theta}_{c}}$
    \STATE$e \gets e+1$    
\ENDWHILE
\STATE \textbf{return} $\boldsymbol{\theta}_{f}$ and $\boldsymbol{\theta}_{c}$
\end{algorithmic}
\end{algorithm}

\section{Experiment and Analysis}
\label{section_experimental_result}
In this section, five different models, namely Multi-task Learning (MTL) \cite{Zhang2022amultitask}, Manifold-based Airfoil Geometric Feature Extraction and Discrepant Data Fusion Learning (MDF) \cite{xiang2023manifold}, RGFil, MLP \cite{xin2022surrogate} and DAN \cite{zuo2023fast} are compared in predicting $C_{P}$ on the wing of DLR-F11 aircraft \cite{coder2015overflow}. Because the DLR-F11 dataset contains both geometry and flight condition parameters of an 3D aircraft, this dataset is chosen for testing all models compared in this paper. 

\subsection{DLR-F11 aircraft dataset and preprocessing}
The DLR-F11 aircraft dataset, provided by the EUROLIFT programme \cite{coder2015overflow}, comprises three parts: geometry part, flight condition part and aerodynamic coefficients. 

\begin{figure}
    \centerline{
        \includegraphics[width=18pc]{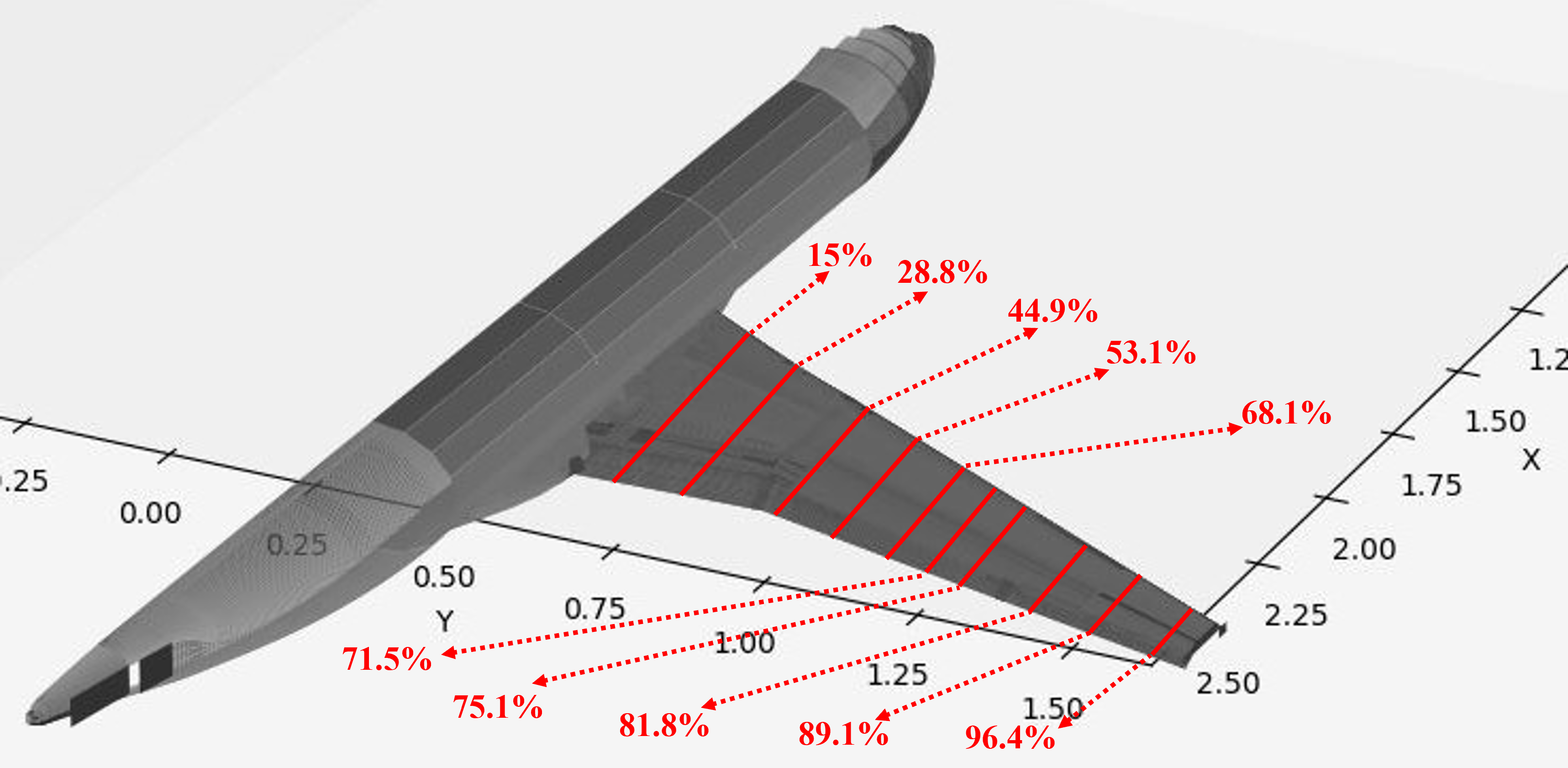}
        }
\caption{The smooth segmented manifold $\mathcal{M}$ and span width configurations.}
\label{fig_f11}
\end{figure}

The geometry part provides the shape of a half fuselage with a wing in the form of polynomial surface control points in 3D space. By (\ref{equ_bezier}), a smooth manifold is constructed based on a part of DLR-F11 control points. Multiple smooth manifolds can be concatenated as a piecewise smooth manifold $\mathcal{M}$ to represent the entire geometry of DLR-f11 aircraft. The geometry part also provides coordinate points with 10 different span width configurations for $C_P$ measurement. The piecewise smooth manifold $\mathcal{M}$ and the span width configurations are shown in Fig. \ref{fig_f11}. For arbitrary point $P_{ab} \in \mathcal{M}$, geometric features ($g_{ij}$, $\mit\Gamma_{i j}^k$ and $S$) can be calculated by (\ref{equ_geometric_feature}), (\ref{equ_gamma}) and (\ref{equ_curvature_ceofficient}), respectively. The dimensions of these geometric features are shown as in Table \ref{tab_Geometric features}.

\begin{table}[h!]
    \centering
    \caption{Dimensions of geometric features.}
    \label{tab_Geometric features}
    \begin{tabular}{cccc}
        \hline
            parameter  & Riemannian metric    &connection          & curvature  \\
        \hline
            dimension  & matrix($2\times 2$)  &matrix($2\times 2\times 2$)    & scalar(1) \\          
        \hline
    \end{tabular}
\end{table}

The flight condition part describes different flight status of DLR-F11 aircraft, the parameters and their variations are shown as in Table \ref{tab_flight_condition}. The aerodynamic coefficient is $C_{P}$.

\begin{table}
\centering
\caption{Flight conditions in DLR-F11 aircraft dataset.}
\label{tab_flight_condition}
\begin{tabular}{cccc}
    \hline
    parameters      & $AoA$                               &   $Ma$    &$Re$             \\
    \hline
    optional value  & \tabincell{c}{0°,7°,12°,16°,18°,\\ 18.5°,19°,20°,21°} & 0.175     &$1.35 \times 10^6$\\  
    \hline
\end{tabular}
\label{tab1}
\end{table}

\subsection{Model settings}
In this section, we introduce the structure and the input data of MTL, MDF, RGFiL, MLP and DAN. 

\begin{table}
\centering
\caption{The optimal structure of comparative models. The symbol ``k'' denotes ``kernel'', and ``s'' denotes ``stride''.}
\label{tab_model_structure}
\begin{tabular}{cccc}
    \hline
    model      & input data                                                       &the optimal structure \\
    \hline
    MTL     & $\vec{\mathbf{x}}_1,\vec{\mathbf{x}}_2$                                               & \tabincell{c}{$f_1$:FC:$16 \times 3$\\
                                                                                                 $f_2$:FC:$16 \times 3$\\
                                                                                                 $c$:FC:$16 \times 3$}        \\ 
    \hline
    MDF     & \tabincell{c}{$\vec{\mathbf{x}}_1,\vec{\mathbf{x}}_2,\vec{\mathbf{x}}_3,$\\
                            $\vec{\mathbf{x}}_4,\vec{\mathbf{x}}_5$}     & \tabincell{c}{$f_1$:FC:$16 \times 3$\\
                                                                                                 $f_2$:FC:$16 \times 3$\\
                                                                                                 $f_3$:\tabincell{c}{CV:k=$2 \times 2 \times 4$,s=$2 \times 2$;\\
                                                                                                                     CV:k=$2 \times 2 \times 8$,s=$2 \times 2$;\\
                                                                                                                     CV:k=$2 \times 2 \times 16$,s=$2 \times 2$;}\\
                                                                                                 $f_4$:\tabincell{c}{CV:k=$2 \times 2 \times 4$,s=$2 \times 2$;\\
                                                                                                                     CV:k=$2 \times 2\times 8$,s=$2 \times 2$;\\
                                                                                                                     CV:k=$2 \times 2 \times 16$,s=$2 \times 2$;}\\
                                                                                                 $f_5$:FC:$16 \times 3$\\
                                                                                                 $c$:FC:$32 \times 3$}\\ 
    \hline
    RGFiL     & \tabincell{c}{$9 \times \{\vec{\mathbf{x}}_1, \vec{\mathbf{x}}_2, $\\
                                    $\vec{\mathbf{x}}_3, \vec{\mathbf{x}}_4, \vec{\mathbf{x}}_5\}$}   & \tabincell{c}{$f_1$:FC:$16 \times 3$\\
                                                                                                 $f_2$:\tabincell{c}{CV:k=$2 \times 2 \times 4$,s=$2 \times 2$;\\
                                                                                                                     CV:k=$2 \times 2 \times 8$,s=$2 \times 2$;\\
                                                                                                                     CV:k=$2 \times 2 \times 16$,s=$2 \times 2$;}\\
                                                                                                 $f_3$:\tabincell{c}{CV:k=$2 \times 2 \times 4$,s=$2 \times 2$;\\
                                                                                                                     CV:k=$2 \times 2 \times 8$,s=$2 \times 2$;\\
                                                                                                                     CV:k=$2 \times 2 \times 16$,s=$2 \times 2$;}\\
                                                                                                 $f_4$:\tabincell{c}{CV:k=$2 \times 2 \times 4$,s=$2 \times 2$;\\
                                                                                                                     CV:k=$2 \times 2 \times 8$,s=$2 \times 2$;\\
                                                                                                                     CV:k=$2 \times 2 \times 16$,s=$2 \times 2$;}\\
                                                                                                 $f_5$:FC:$16 \times 3$\\
                                                                                                 $c$:FC:$16\times 3$}\\
    \hline 
    MLP     & \tabincell{c}{$9 \times \{\vec{\mathbf{x}}_1, \vec{\mathbf{x}}_2,$ \\
                           $\vec{\mathbf{x}}_3, \vec{\mathbf{x}}_4, \vec{\mathbf{x}}_5\}$}                               & FC:$128 \times 6$\\
    \hline
    DAN        &  \tabincell{c}{wing section image,\\
                                 $\vec{\mathbf{x}}_1$}                                                  & FC:$128 \times 6$           \\
    \hline
\end{tabular}
\label{tab1}
\end{table}

Flight conditions, coordinates on the wing surface, and geometric features are different types of data, and their numerical value, distribution, and physical meaning are different. It has been demonstrated that using a context network to learn the weight vector to fuse $f_{z\alpha}(\vec{\mathbf{x}}_z)$ outputs more accurate predicted coefficients than with directly concatenates $\vec{\mathbf{x}}_1 \sim \vec{\mathbf{x}}_5$ as inputs \cite{xiang2023manifold, deng2023prediction, wang2023airfoil}. In this paper, MTL, MDF and RGFiL use a weight vector mechanism to predict $C_{P}$, MLP concatenates $\vec{\mathbf{x}}_1 \sim \vec{\mathbf{x}}_5$ as inputs to predict $C_{P}$, and DAN concatenates features extracted from wing section images and flight conditions together as inputs to predict $C_{P}$.

Considering that the above models contain fully connected (FC) layers and convolutional (CV) layers, we design multiple structures for each model according to the number of FC layers, CV layers and hidden nodes. The optional numbers of CV and FC layers are 3 and 6, respectively. For FC layers, the optional numbers of hidden nodes are 16, 32, 64 and 128. For CV layers, the kernel size are $2 \times 2$, the channel of each kernel is $2^{1+e}$, where $e$ denotes the sequence number of each CV layer ($e$ starts from 1). For MTL, MDF and RGFiL, the number of hidden layers in each function network are the same (for example, if the function network\_1 has 3 layers, the remaining function networks also have 3 hidden layers). Therefore, MTL, MDF and RGFiL have 64 different structures, respectively. MLP and DAN have 8 different structures, respectively. Table \ref{tab_model_structure} only introduce the structures with smallest test errors (the optimal structure).   

MTL takes flight conditions $\vec{\mathbf{x}}_1$ and coordinates $\vec{\mathbf{x}}_2$ as inputs to predict $C_{P}$. The function network\_1, taking $\vec{\mathbf{x}}_1$ as input, has 3 hidden layers, each of which has 16 nodes, i.e., the structure of function network\_1 is set to ``FC: $16 \times 3$''. The function network\_2 takes $\vec{\mathbf{x}}_2$ (with shape $batchsize \times 3$) as input, the structure of function network\_2 is set to ``FC: $16 \times 3$''. In addition, the context network takes $[\vec{\mathbf{x}}_1,\vec{\mathbf{x}}_2]$ as input, and the structure of the context network is set to ``FC: $16 \times 3$''. 

MDF in \cite{xiang2023manifold} takes $\vec{\mathbf{x}}_1$, $\vec{\mathbf{x}}_2$ and $\vec{\mathbf{x}}_3$ as inputs to predict $C_{L}$ and $C_{D}$.  In this experiment, MDF takes $\vec{\mathbf{x}}_1$, $\vec{\mathbf{x}}_2$ and geometric features ($\vec{\mathbf{x}}_3$, $\vec{\mathbf{x}}_4$, and $\vec{\mathbf{x}}_5$) as inputs to predict $C_{P}$. The function network\_1 takes $\vec{\mathbf{x}}_1$ as input, and the function network\_2 takes $\vec{\mathbf{x}}_2$ as input. The structures of function network\_1 and function network\_2 in MDF are identical to those of function network\_1 and function network\_2 in MTL. The function network\_3 that takes $\vec{\mathbf{x}}_3$ (with shape $batchsize \times 1 \times 2 \times 2$) as input is designed as a 2D CNN. The function network\_3 consists of three convolutional layers, each of which have a convolution kernel with shape $2 \times 2 \times 4$, $2 \times 2 \times 8$ and $2 \times 2 \times 16$, respectively. The function network\_4 takes $\vec{\mathbf{x}}_4$ (with shape $batchsize \times 2 \times 2\times 2$) as input. The structure of function network\_4 is the same with the structure of function network\_3. The stride of all kernels is $2 \times 2$. Because $\vec{\mathbf{x}}_5$ is a vector with shape $batchsize \times 1$, the structure of the function network\_5 is set to ``FC: $16 \times 3$''. In addition, the structure of the context network is set to ``FC: $32 \times 3$''. 

\begin{figure}
\centerline{
        \includegraphics[width=16pc]{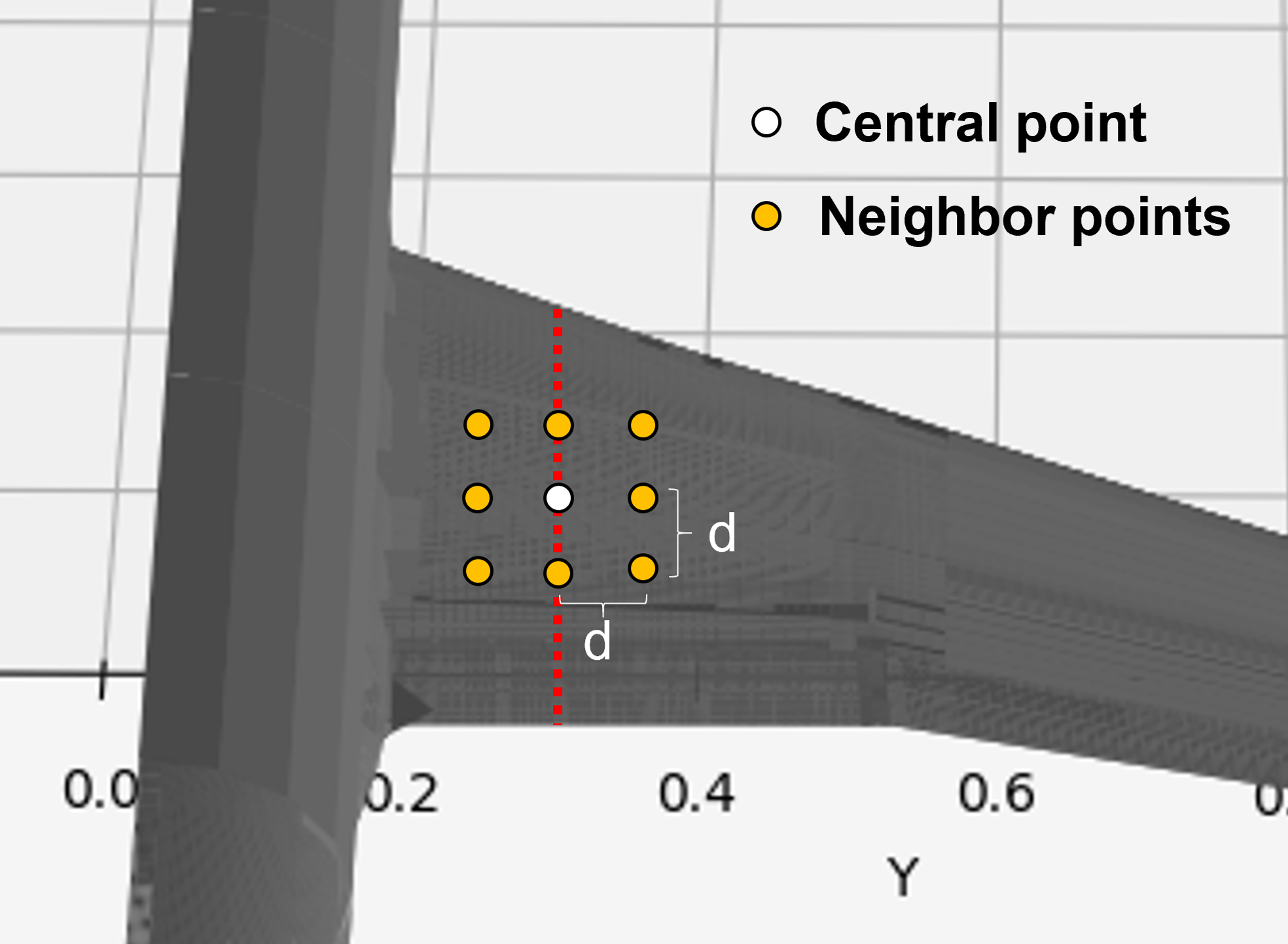}
        }
\caption{Vertical and horizontal Euclidean distance among neighbors.}
\label{fig_neighbors}
\end{figure}

RGFiL additionally considers geometric features from 8 neighbors around point $P_{ab}$. Fig. \ref{fig_neighbors} describes how to select the 8 neighbors on the DLR-F11 wing. The distance $d$ is a hyperparameter with optional values of 0.01, 0.005, and 0.001. The input of function network\_2 is a matrix (with shape $batchsize \times 1 \times 9 \times 3$) that contains 9 different $\vec{\mathbf{x}}_2$, therefore the function network\_2 is a 2D CNN, which consists of three convolutional layers, each of which have a convolution kernel with shape $2 \times 2 \times 4$, $2 \times 2 \times 8$ and $2 \times 2 \times 16$, respectively. Similarly, the inputs of the function network\_3 (with shape $batchsize \times 1 \times 18\times 2$), function network\_4 (with shape $batchsize \times 2 \times 18\times 2$) and function network\_5 (with shape $batchsize \times 9$) are also matrices contain 9 different $\vec{\mathbf{x}}_3$, $\vec{\mathbf{x}}_4$ and $\vec{\mathbf{x}}_5$, respectively. The structures of function network\_1, function network\_5 and context network are all set to ``FC: $16 \times 3$''.

The input data of MLP is exactly the same with RGFiL. The MLP is a multi-layer perceptron with 6 layers and each layer has 128 nodes. The aim of comparing MLP with MTL, MDF and RGFiL is to further verify the effectiveness of the weight vector mechanism. Because the source code of MLP is not published by \cite{xin2022surrogate}, we implemented MLP for this experiment.

DAN takes wing section images with different span width and $\vec{\mathbf{x}}_1$ as input to predict $C_{P}$. DAN consists of two parts, a pre-trained Transformer encoder and a MLP. The pre-trained transformer encoder is used to extract latent features (with shape $batchsize \times 1000$) from airfoil images. The input of the MLP in DAN is the concatenation of extracted features and $\vec{\mathbf{x}}_1$. The optimal MLP in DAN  has 6 layers and each of which has 128 nodes \footnote{The source code of DAN is downloaded from https://github.com/zuokuijun/vitAirfoilEncoder.}.

\begin{table}
    \centering
    \caption{Parameters of all models in this paper.}
    \label{tab_model_parameter}
    \begin{tabular}{cc}
        \hline
        parameter                     &value                    \tabularnewline
        \hline
        optimization algorithm              &Adam(Beta1=0.9 and Beta2=0.999)\cite{yamin2022neural} \tabularnewline
        data normalization                  &max-min normalization    \tabularnewline
        activation function                 &LeakyReLU \cite{ivan2023empirical} \tabularnewline
        batch size                          &470                      \tabularnewline
        learning rate                       &0.001                   \tabularnewline
        epochs                              &2000                     \tabularnewline
        validation                          &7-fold cross validation  \tabularnewline
        \hline
    \end{tabular}
\end{table}

\begin{table*}[t]
\centering
\caption{Average MSEs of MTL\cite{Zhang2022amultitask}, MDF\cite{xiang2023manifold}, RGFiL, MLP\cite{xin2022surrogate} and DAN\cite{zuo2023fast}. $\eta$ denotes the MSE reduction of RGFiL\_2 compared with DAN. The average MSE reduction is calculated as the average value for each MSE reduction.  }
\label{tab_errors}
  \begin{tabular}{ccccccccccc}
    \hline
    $AoA$& MTL & MDF & \tabincell{c}{RGFiL\_1\\($d=0.01$)} &\tabincell{c}{RGFiL\_2\\($d=0.005$)} &\tabincell{c}{RGFiL\_3\\($d=0.001$)} & \tabincell{c}{MLP\_1\\($d=0.01$)} &\tabincell{c}{MLP\_2\\($d=0.005$)}&\tabincell{c}{MLP\_3\\($d=0.001$)}& DAN & $\eta$\\ 
    \hline
    7°   & 1.49    & 1.53    & \textbf{1.46}   &1.47   &1.49   & 1.53   &1.52   &1.54    & 1.52   &3.28\%\\
    12°  & 1.19    & 1.21    & 1.23   &\textbf{1.15}   &1.18   & 1.21   &1.20   &1.20    & 1.21   &4.95\%\\
    16°  & 8.26E-1 & 6.66E-1 & 6.54E-1&5.97E-1&6.44E-1&\textbf{5.93E-1}&6.28E-1&6.18E-1 & 6.50E-1&8.15\%\\
    18°  & 3.31E-1 & 2.92E-1 & 2.03E-1&\textbf{2.01E-1}&2.24E-1& 2.22E-1&2.26E-1&2.17E-1 & 2.29E-1&12.23\%\\
    18.5°& 2.22E-1 & 1.82E-1 & 1.50E-1&\textbf{1.21E-1}&1.86E-1& 1.36E-1&1.53E-1&1.62E-1 & 1.56E-1&22.43\%\\
    19°  & 1.71E-1 & 2.08E-1 & 7.72E-2&7.28E-2&\textbf{6.87E-2}& 1.99E-1&8.42E-2&9.21E-2 & 7.98E-2&8.77\%\\
    20°  & 1.70E-1 & 6.94E-2 & 1.41E-2&\textbf{1.26E-2}&2.66E-2& 1.82E-2&1.73E-2&2.97E-2 & 2.30E-2&45.22\%\\
    average&6.29E-1& 5.94E-1 & 5.41E-1&\textbf{5.18E-1}&5.46E-1& 5.58E-1&5.47E-1 &5.51E-1& 5.53E-1&15.00\%\\
    \hline
  \end{tabular}
\end{table*}

The parameters of all models compared in this paper are shown in Table \ref{tab_model_parameter}. The MSE are chosen as an assessment indicator for the predicted $C_{P}$:
\begin{equation}
\label{equ_mse_mae_performance}
\nonumber
  MSE =\frac{1}{M} \sum_{\beta=1}^{M}\left(C_{P_{te}}^\beta-\hat{C}_P^\beta\right)^{2}
\end{equation}
where $C_{P_{te}}^\beta$ denotes the true $C_P$ value of the $\beta$th sample in the test set, and $\hat{C}_P^\beta$ denotes the predicted value of the $\beta$th sample in the test set.

\begin{figure*}[t]
    \centering
    \subfigure[]{
       \includegraphics[scale=0.3]{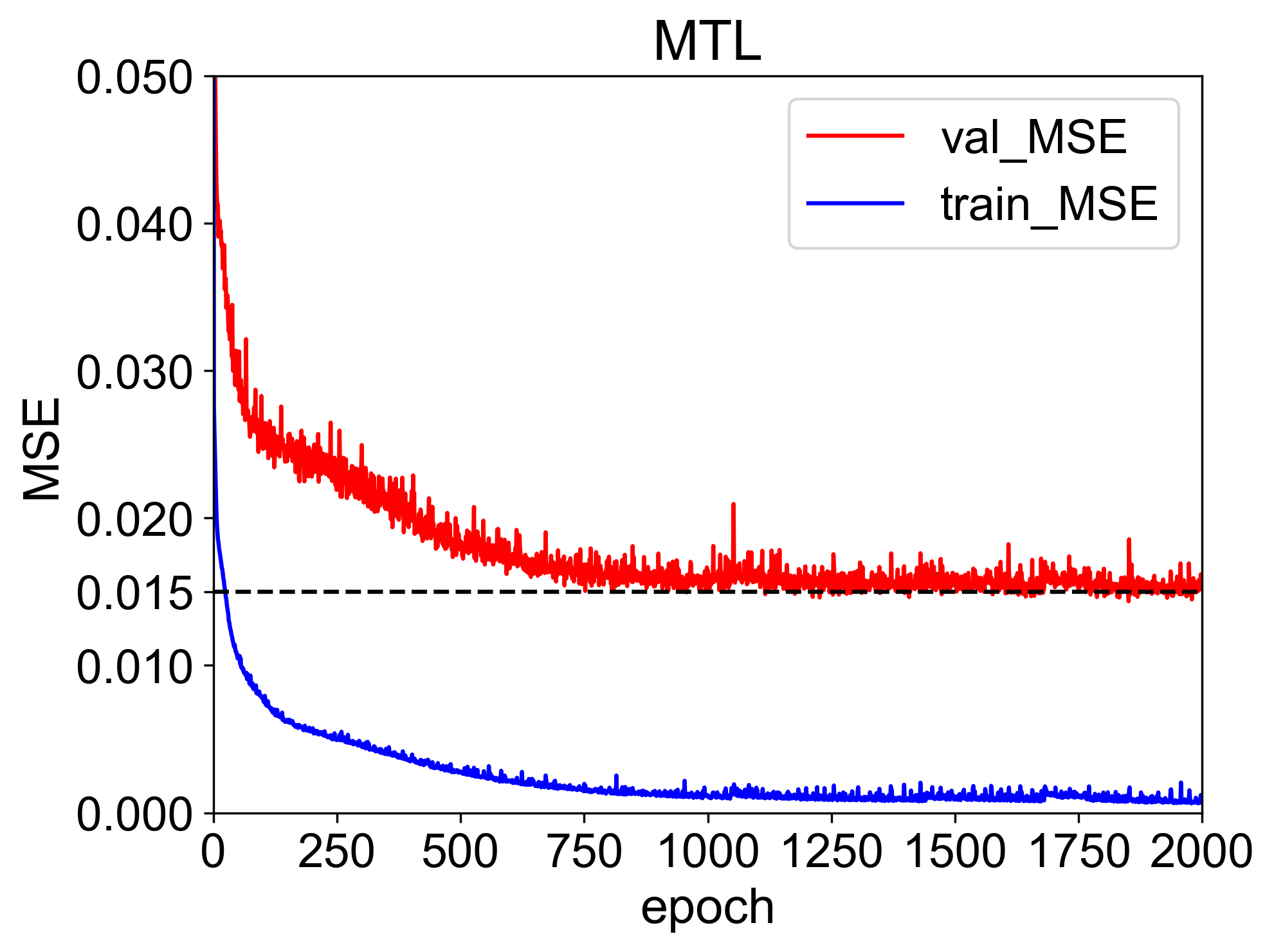}
    }
    \subfigure[]{
       \includegraphics[scale=0.3]{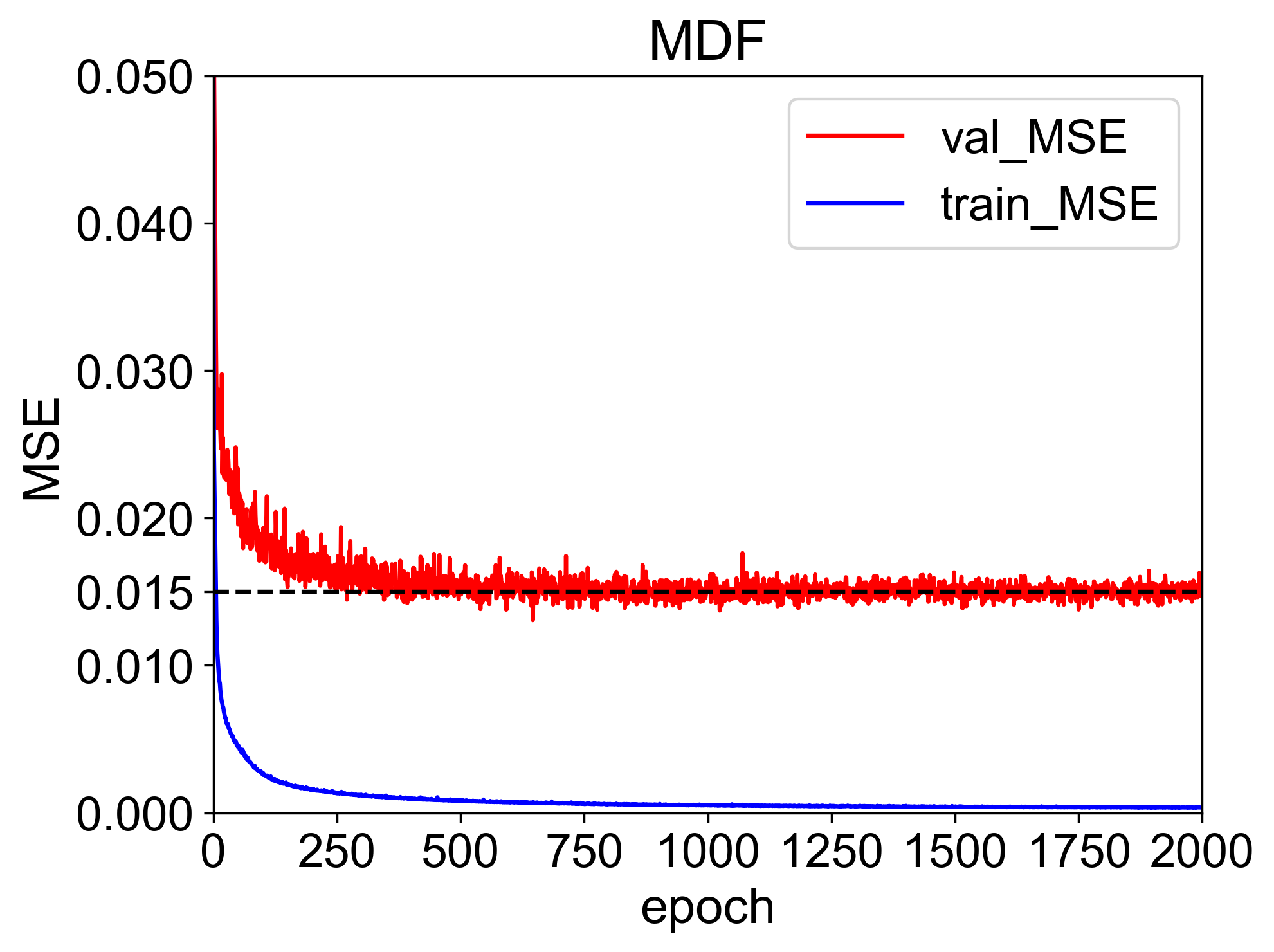}
    }
    \subfigure[]{
       \includegraphics[scale=0.3]{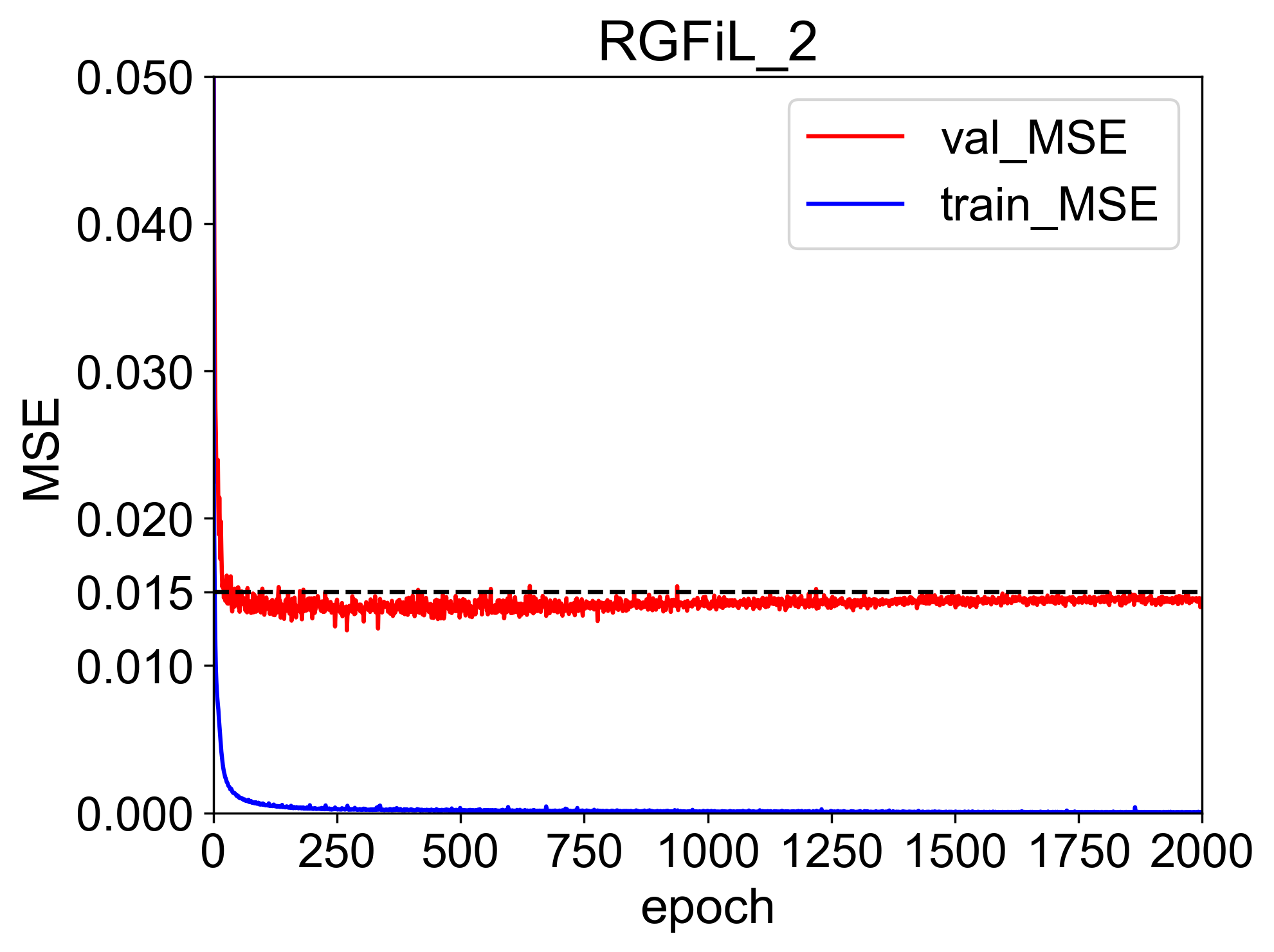}
    }\\
    \subfigure[]{
       \includegraphics[scale=0.3]{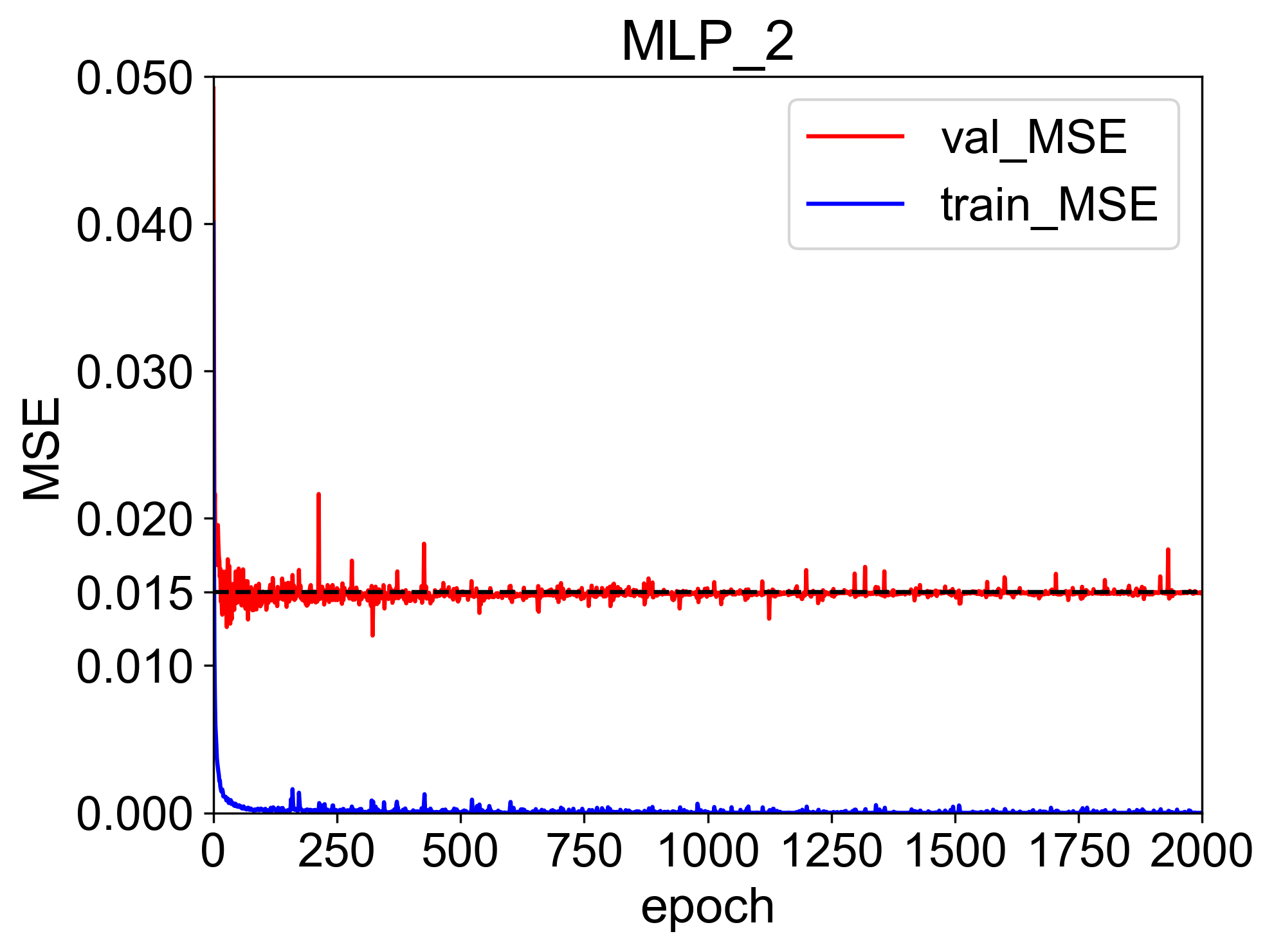}
    }
    \subfigure[]{
       \includegraphics[scale=0.3]{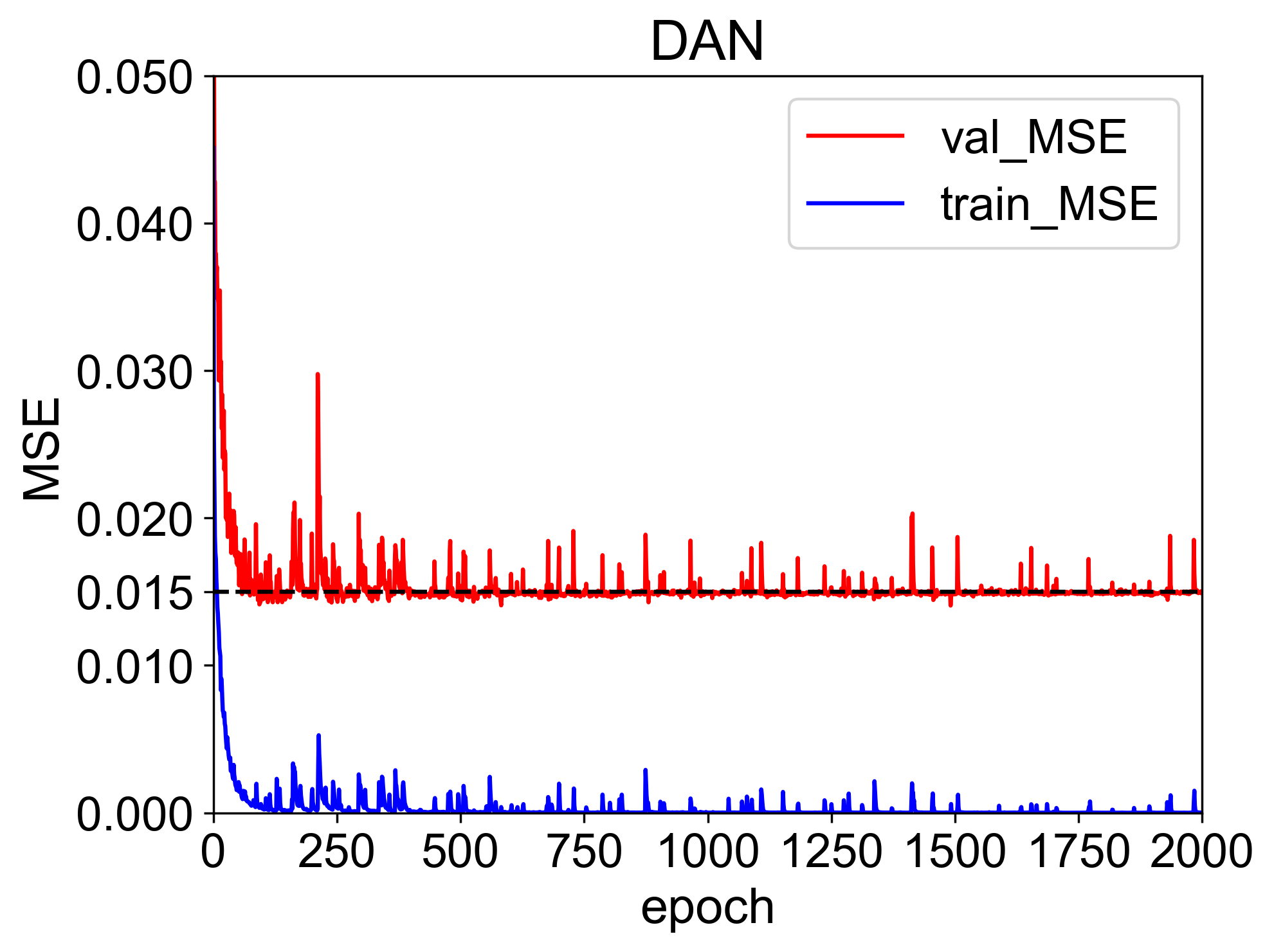}
    }
    \caption{Average training and validation loss curves for MTL, MDF, RGFiL\_2, MLP\_2 and DAN.}
    \label{fig_loss}
\end{figure*}

All models in this paper are verified by a 7-fold cross-validation. The test set of each fold experiment in the 7-fold cross-validation contains all the samples with a specific $AoA$, and the samples related to this $AoA$ do not appear in the training set. For example, in the first-fold of the 7-fold cross-validation, all the data samples with $AoA = 7$° are selected as the first test set, the remaining sample are selected as the first training set. In the second-fold of the 7-fold cross-validation, all the samples with $AoA = 12$° are selected as the second test set, the remaining sample are selected as the second training set. The remaining folds of the 7-fold cross-validation are similar.

\subsection{Results and analysis}

\begin{figure*}
    \centering
    \subfigure[]{
       \includegraphics[scale=0.45]{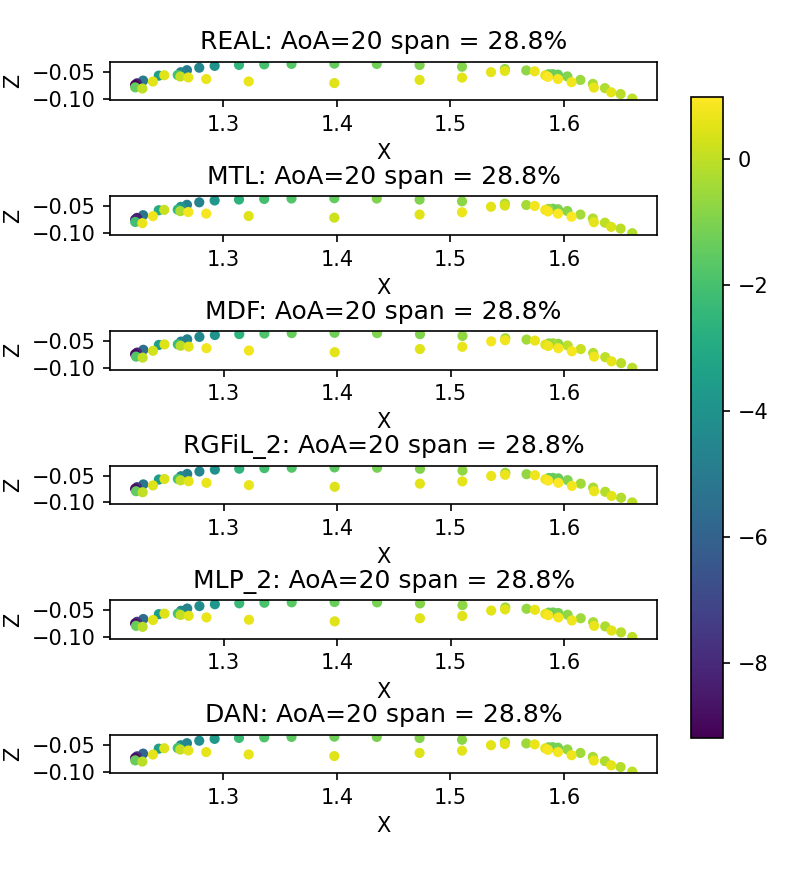}
    }
    \quad
    \subfigure[]{
       \includegraphics[scale=0.45]{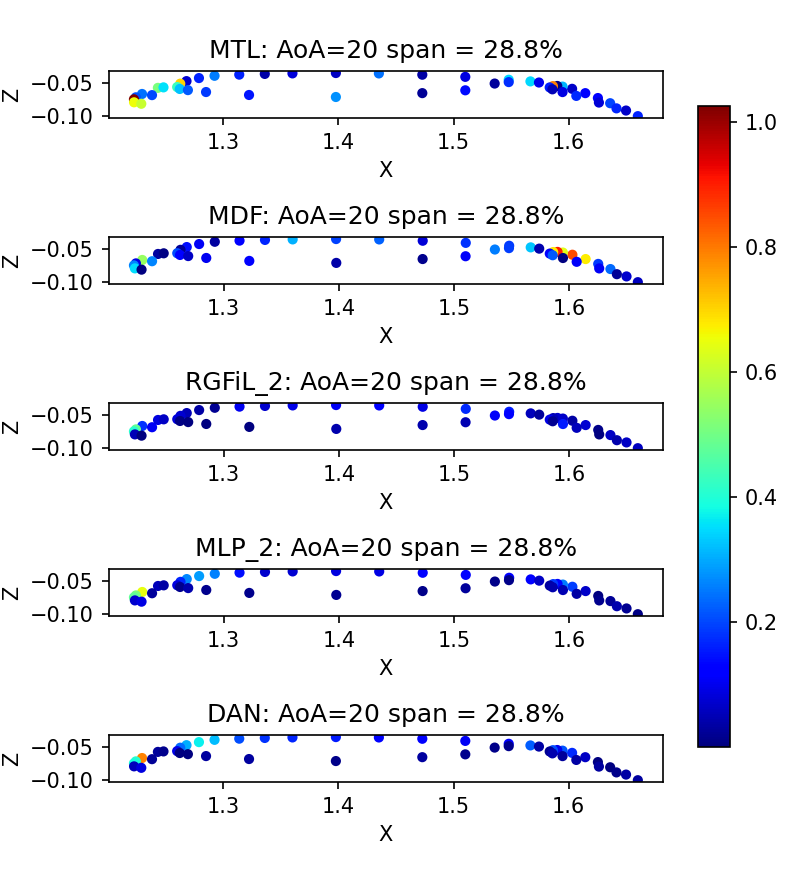}
    }
    \caption{The predicted $C_P$ (a) and predicted errors (b) with $AoA=20$° and span width equals 28.8\%. The horizontal axis denotes the x axis along the fuselage direction, and the vertical axis denotes the z axis along the fuselage normal direction.}
    \label{fig_cp_and_cha_20_288}
\end{figure*}

Table \ref{tab_errors} displays the average test MSEs for MTL, MDF, RGFiL, MLP and DAN. In every row, figures in bold represent the minimum test MSEs (all the test MSEs are not normalized). Generally, the test MSEs of $C_{P}$ predicted by all models decrease as the $AoA$ increases. This phenomenon occurs because the data samples in the large $AoA$ region are densely populated, whereas the data samples in the small $AoA$ region are relatively sparse. Among RGFiL\_1, RGFiL\_2 and RGFiL\_3, the average test MSEs of RGFiL\_2 is smallest. This indicates that RGFiL show smaller prediction MSEs with $d=0.005$. Comparing RGFiL with MLP, RGFiL show smaller average test MSEs than MLP, which indicates the advantage of the multi-feature learning structure. Comparing RGFiL\_2 with DAN, RGFiL\_2 also exhibits smaller test MSEs than DAN. In addition, RGFiL\_2 reduces the test MSEs of $C_P$ by an average of 15.00\%, compared with DAN. The above observations indicate that RGFiL\_2 is more accurate in predicting $C_P$ than other models.

\begin{figure*}
    \centering
    \subfigure[]{
       \includegraphics[scale=0.45]{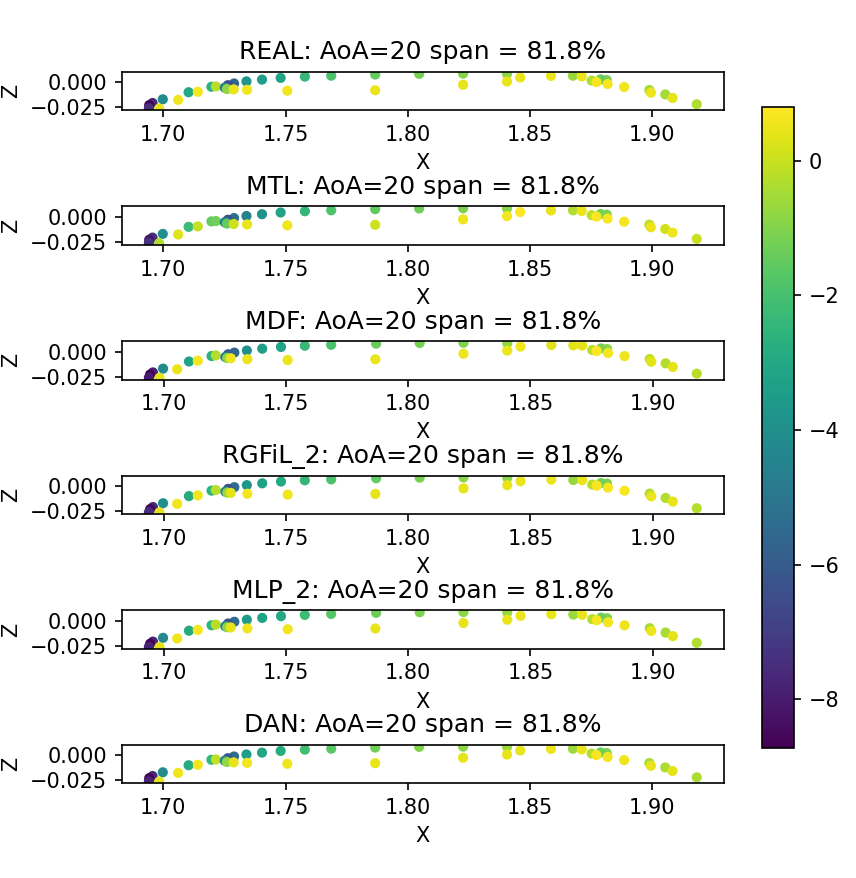}
    }
    \quad
    \subfigure[]{
       \includegraphics[scale=0.45]{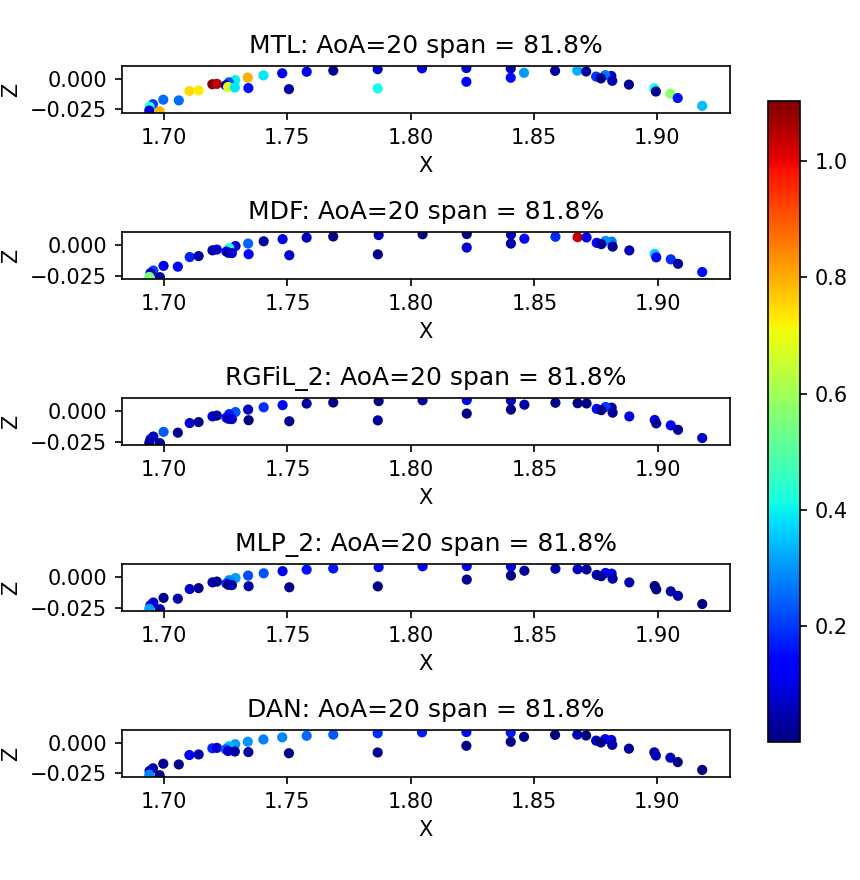}
    }
    \caption{The predicted $C_P$ (a) and predicted errors (b) with $AoA=20$° and span width equals 81.8\%. The horizontal and vertical axes have the same meanings as depicted in Fig. \ref{fig_cp_and_cha_20_288}.}
    \label{fig_cp_and_cha_20_818}
\end{figure*}

Fig. \ref{fig_loss} depicts the average training and validation curves for all the above models in the 7-fold experiments. The fluctuations in loss variations observed in subgraphs (a), (b), (d) and (e) are larger than those in subgraph (c). This indicates a difference in the stability of the training processes (i.e., the training progress of RGFiL\_2 is more stable). Furthermore, compared with subgraph (a), the models represented in subgraphs (b), (c) and (d) (MDF, RGFiL\_2 and MLP\_2) converge faster than MTL. These three models all incorporate geometric features as part of their inputs, which suggests that the incorporation of geometric features accelerates the convergence of their training processes. It's worth highlighting that the models corresponding to subgraphs (c) and (d) (RGFiL\_2 and MLP\_2) incorporate the geometric features from additional 8 neighbors as part of their inputs, which further accelerates the convergence and further reduce the test MSEs, but at the cost of little overfitting.

\begin{figure*}
    \centering
    \subfigure[]{
       \includegraphics[scale=0.45]{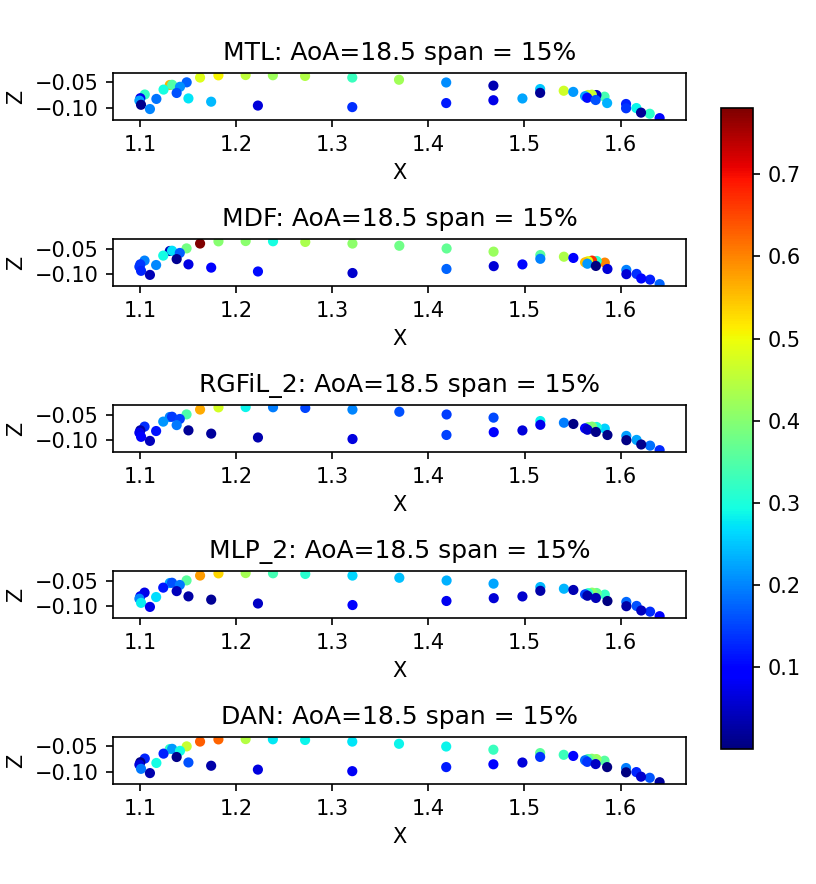}
    }
    \quad
    \subfigure[]{
       \includegraphics[scale=0.45]{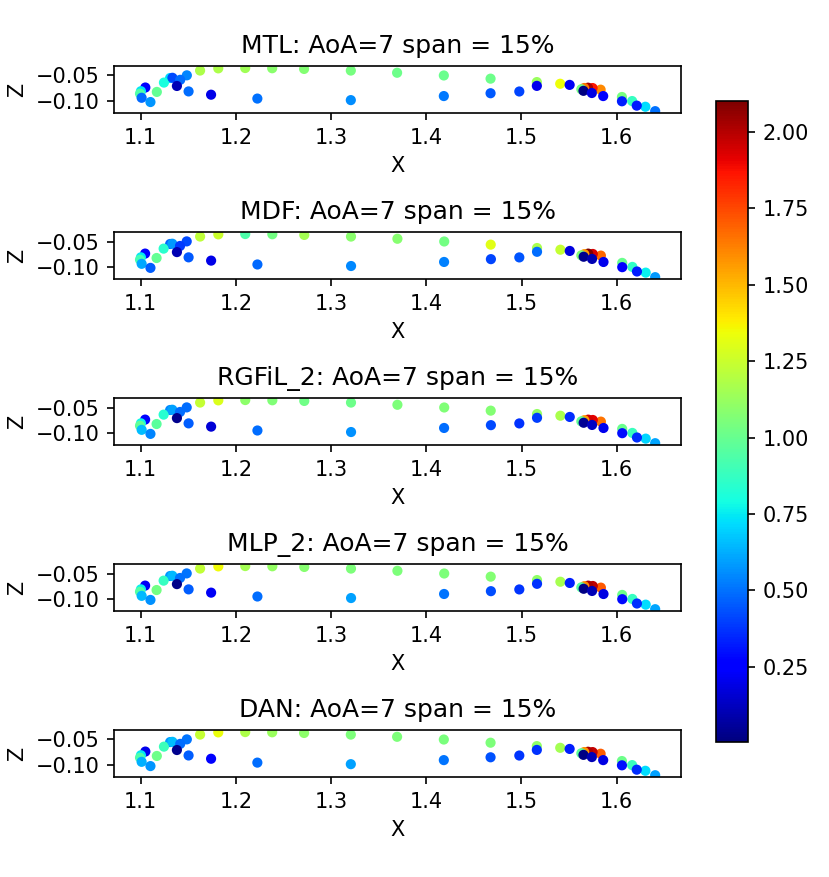}
    }
    \caption{The predicted errors with $AoA=18.5$° (a) and $AoA=7$° (b), and span width equals 15\%. The horizontal and vertical axes have the same meanings as depicted in Fig. \ref{fig_cp_and_cha_20_288}.}
    \label{fig_cp_and_cha_7_185}
\end{figure*}

Fig. \ref{fig_cp_and_cha_20_288} illustrates the predicted $C_{P}$ distributions and the corresponding prediction errors on the surface of DLR-F11 wing with $AoA = 20$° and span width equals 28.8\%. The DLR-F11 aircraft contains a three-stage wing, it is evident that the wing sections depicted in the graph are composed of three parts. In each row of every subgraph, the left part represents the first wing stage, the middle part represents the second wing stage, and the right part denotes the third wing stage. In subgraph (a), the $C_{P}$ distributions predicted by all models are similar, suggesting that each model is capable of learning the variations in $C_{P}$ distribution on the DLR-F11 wing surface. To clearly compare the details of the predicted $C_P$, we calculate

\begin{equation}
\label{equ_cp_cha}
  err^\beta = | C_{P_{te}}^\beta - \hat{C}_P^\beta |
\end{equation}
where $err^\beta$ denotes the prediction error of the $\beta$th sample, $C_{P_{te}}^\beta$ denotes the real $C_P$ value of the $\beta$th sample in the test set, and $\hat{C}_P^\beta$ denotes the predicted value of the $\beta$th sample in the test set. The $err^\beta$ of each model is plotted in subgraph (b). On the first wing stage, most of the points generated by MTL and MDF show larger errors compared to the points generated by RGFiL\_2. At the leading edge on the three wing stages, the points generated by MTL, MDF, MLP\_2 and DNA show larger errors than the points generated by RGFiL\_2. Because the leading edge of a wing is the area where its $C_{P}$ changes dramatically, RGFiL\_2 has an advantage at learning $C_P$ variation at the leading edge of DLR-F11 wing. At the trailing edge on the lower surface of the second wing stage, the points generated by RGFiL\_2 show larger errors than the points generated by MLP\_2 and DAN. These observations demonstrate that RGFiL\_2 is more accurate at the leading edge of wing sections than other methods.

\begin{figure*}
    \centering
    \subfigure[]{
       \includegraphics[scale=0.45]{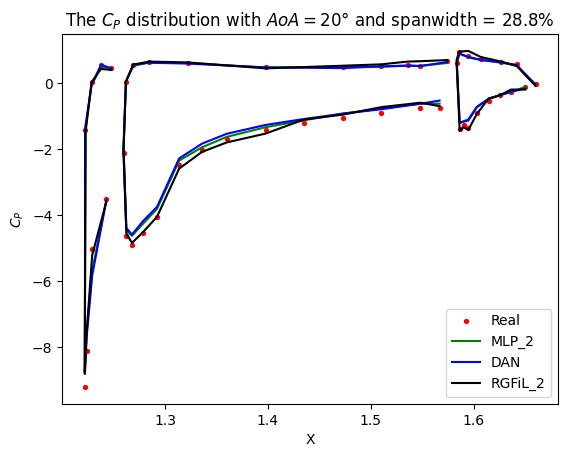}
    }
    \quad
    \subfigure[]{
       \includegraphics[scale=0.45]{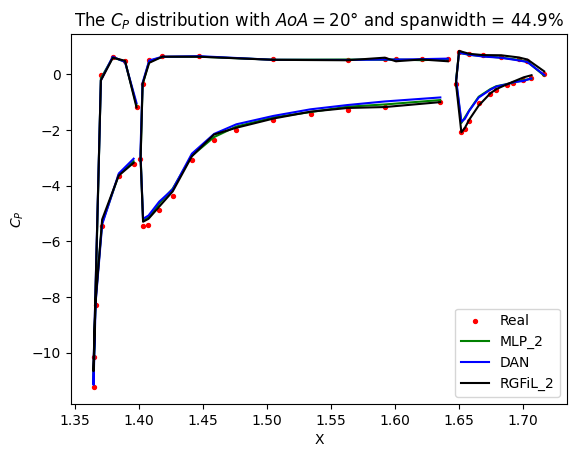}
    }\quad
    \subfigure[]{
       \includegraphics[scale=0.45]{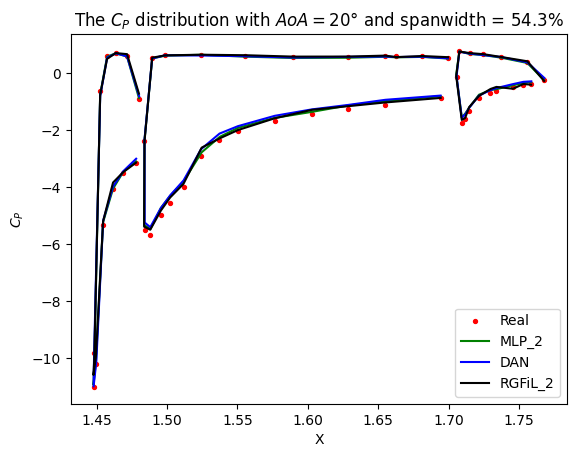}
    }\quad
    \subfigure[]{
       \includegraphics[scale=0.45]{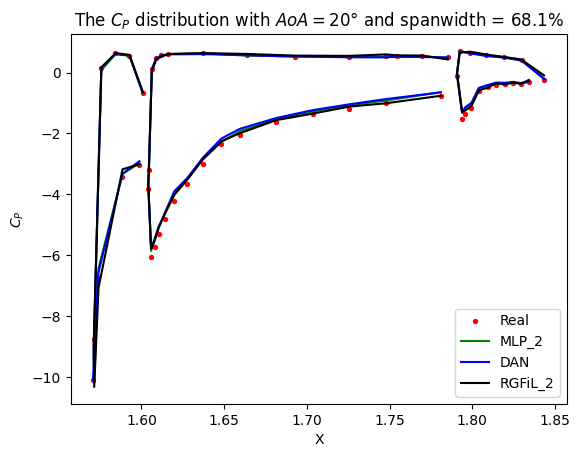}
    }\quad
    \subfigure[]{
       \includegraphics[scale=0.45]{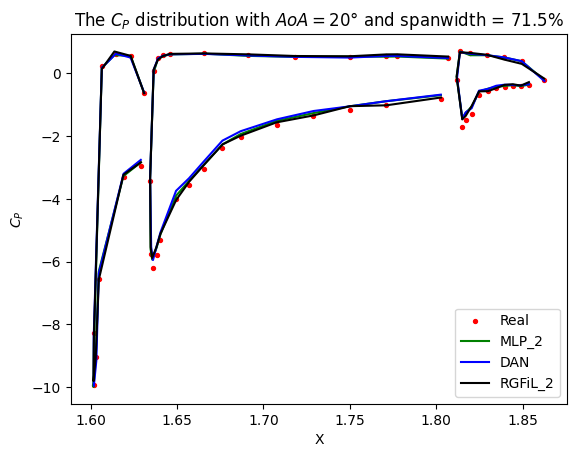}
    }\quad
    \subfigure[]{
       \includegraphics[scale=0.45]{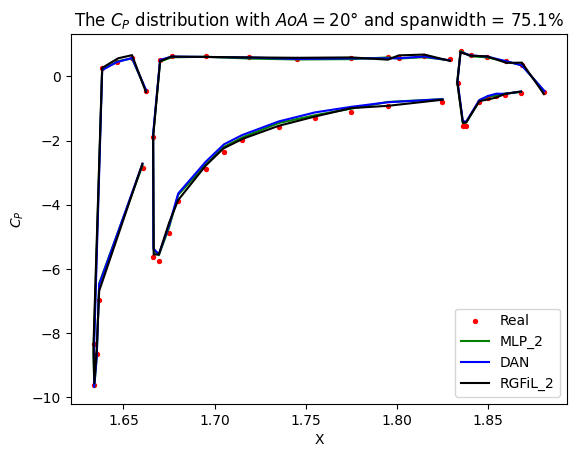}
    }\quad
    \caption{The $C_{P}$ variations of test set (Real), RGFiL\_2, MLP\_2 and DAN with $AoA=20$°. The horizontal axis denotes the x axis along the fuselage direction, and the vertical axis denotes $C_P$ value.}
    \label{fig_cp_variation_20}
\end{figure*}

Fig. \ref{fig_cp_and_cha_20_818} illustrates the predicted $C_{P}$ distributions and the corresponding prediction errors on the surface of DLR-F11 wing with $AoA = 20$° and span width equals 81.8\%. Similar to Fig. \ref{fig_cp_and_cha_20_288}, subgraph (a) demonstrates that all models can learn the variations of $C_{P}$ distributions. In subgraph (b), on the first wing stage, the points generated by MTL and MDF have larger errors than the points generated by RGFiL\_2, MLP\_2 and DAN. At the leading edge on the upper surface of the second wing stage, the points generated by MTL, MDF, MLP\_2 and DAN have larger errors than the points generated by RGFiL\_2. On the third wing stage, the points generated by MTL and MDF show larger errors than the points generated by RGFiL\_2, MLP\_2 and DAN. These observations demonstrate that RGFiL\_2 is more accurate at the leading edge on the second wing stage than other methods.

\begin{figure*}
    \centering
    \subfigure[]{
       \includegraphics[scale=0.45]{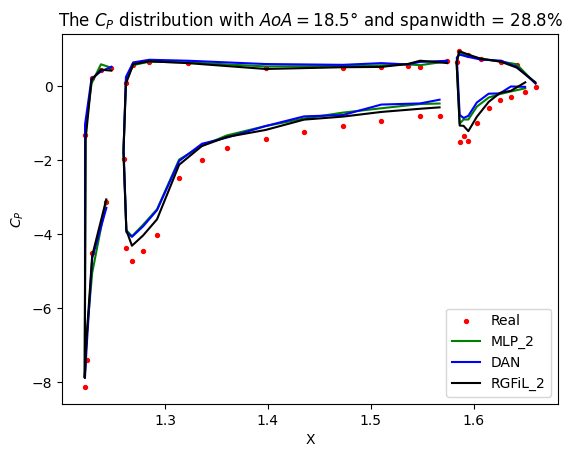}
    }
    \quad
    \subfigure[]{
       \includegraphics[scale=0.45]{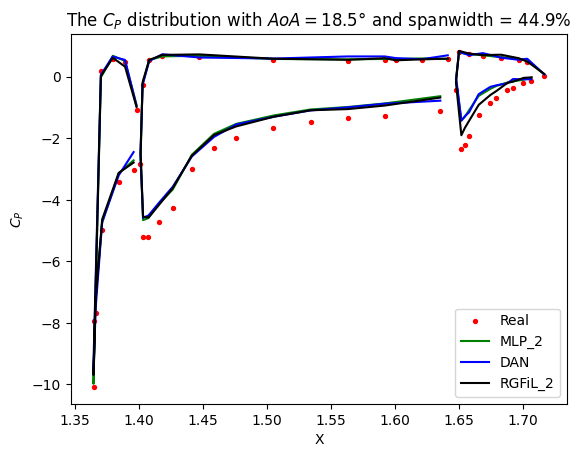}
    }\quad
    \subfigure[]{
       \includegraphics[scale=0.45]{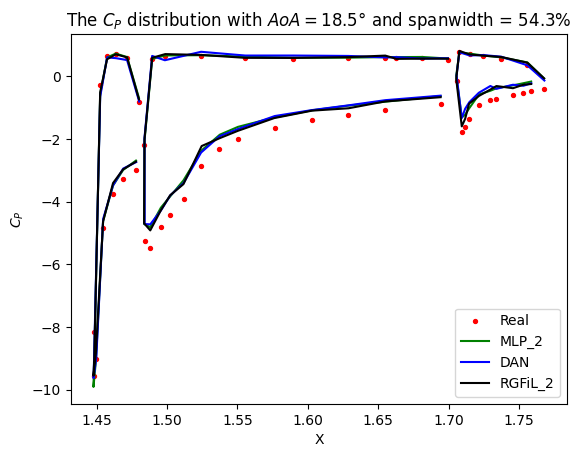}
    }\quad
    \subfigure[]{
       \includegraphics[scale=0.45]{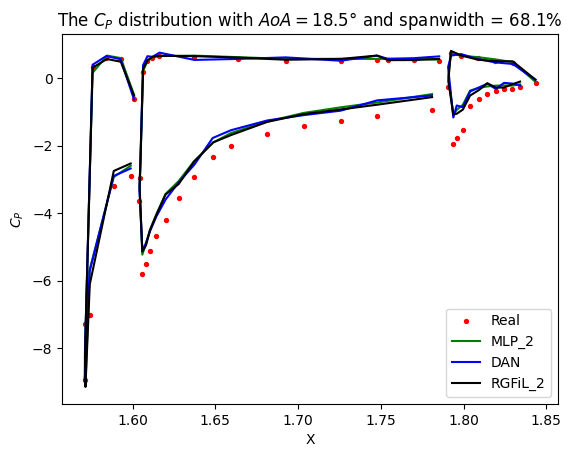}
    }\quad
    \subfigure[]{
       \includegraphics[scale=0.45]{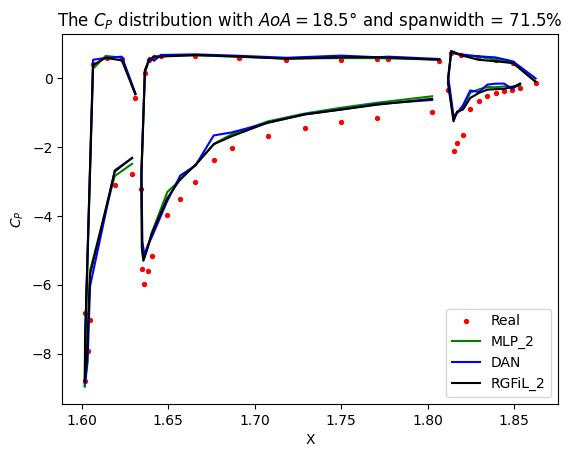}
    }\quad
    \subfigure[]{
       \includegraphics[scale=0.45]{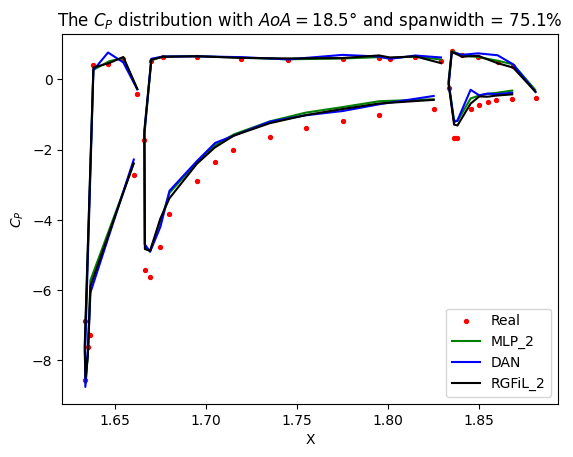}
    }\quad
    \caption{The $C_{P}$ variations of test set (Real), RGFiL\_2, MLP\_2 and DAN with $AoA=18.5$°. The horizontal and vertical axes have the same meanings as depicted in Fig. \ref{fig_cp_variation_20}.}
    \label{fig_cp_variation_18.5}
\end{figure*}

Fig. \ref{fig_cp_and_cha_7_185} displays the prediction errors of the DLR-F11 wing section with a span width of 15\% at $AoA$ of 18.5° and 7°, respectively. In subgraph (a), at the leading edge on the first wing stage, the points generated by MTL, MDF, MLP\_2 and DAN show larger errors than the points generated by RGFiL\_2. On the upper surface of the second stage, the points generated by  MTL, MDF, MLP\_2 and DAN exhibit larger errors than the points generated by RGFiL\_2. On the lower surface of the second stage and both the upper surface and lower surface of the third stage, the points generated by RGFiL\_2, MLP\_2 and DAN are similar. In subgraph (b), it is difficult to see the differences among the models in detail. In addition, points on the upper surface of the second wing stage generated by all methods have large errors, because the DLR-F11 data
samples in small $AoA$ region are sparser than those in large $AoA$. The observations in Fig. \ref{fig_cp_and_cha_20_288}, Fig. \ref{fig_cp_and_cha_20_818} and Fig. \ref{fig_cp_and_cha_7_185} demonstrate that RGFiL\_2 exhibits distinct advantages at the leading edge of wing sections in case with large $AoA$, while in cases with small $AoA$, RGFiL\_2 closely aligns with existing methods, which is consistent with table \ref{tab_errors}.

To better illustrate $C_{P}$ variations of different models, the $C_{P}$ variations of different wing sections at different $AoA$ are plotted in Fig. \ref{fig_cp_variation_20} and Fig. \ref{fig_cp_variation_18.5}. It is worth noting that only three models (RGFiL\_2, MLP\_2 and DAN) are included to maintain clarity in graphs, as they exhibit relatively small prediction errors. In each subgraph, three set of curves are plotted to depict the $C_{P}$ variations on three stages of the DLR-F11 wing. The left curve represents the $C_{P}$ variation on the first wing stage, the middle curve represents the $C_{P}$ variation on the second wing stage, and the right curve represents the $C_{P}$ variation on the third wing stage. In each curve, the upper curve represents the $C_{P}$ variations on the lower surface of the DLR-F11 wing, and the lower curve represents the $C_{P}$ variations on the upper surface of the DLR-F11 wing.

Fig. \ref{fig_cp_variation_20} displays the $C_{P}$ variations of multiple wing sections with span width of 28.8\%, 44.9\%, 54.3\%, 68.1\%, 71.5\% and 75.1\% at $AoA$ of 20°. In subgraphs (a) and (b), the black curves closely align with the red points in the lower part of each curve, indicating that RGFiL\_2 effectively captures the $C_{P}$ variations on the upper surface of the DLR-F11 wing. In subgraphs (c) and (d), all curves exhibit similar variations. In subgraphs (e) and (f), the black curves at the first and third wing stage closely align with the red points. The black lower curve of the second wing stage are closer to the red points than other curves. In general, RGFiL\_2 accurately predicts $C_{P}$ variations on the three wing stages, especially on the upper surface of the second wing stage. However, RGFiL\_2 are inaccurate in predicting $C_P$ value of few points on the lower surface of the second wing stage (Notice: the upper curve in Fig. \ref{fig_cp_variation_20} denotes the $C_{P}$ variations on the lower surface of the DLR-F11 wing, and the lower curve denotes the $C_{P}$ variations on the upper surface of the DLR-F11 wing).

Fig. \ref{fig_cp_variation_18.5} illustrates $C_{P}$ variations of the same wing sections with the same span widths at $AoA = 18.5$°. In all subgraphs, the lower part of the blue curves and green curves on the third wing stage deviate from the red points, and the blue curves deviate more significantly. In contrast, the black curves in the corresponding region are closer to the red points. This observation indicates that DAN and MLP\_2 are difficult to accurately predict the $C_{P}$ variation on the upper surface of the third wing stage, while RGFiL\_2 can predict more accurate $C_{P}$ variations than DAN and MLP\_2 in this region. In subgraphs (a), (d), (e) and (f), the lower part of the black curves on the second wing stage are closer to the red points than the corresponding part of both the blue curves and green curves. These observations suggest that RGFiL\_2 proposed in this paper offer a more accurate representation of 3D wings than DAN and MLP\_2.

\begin{figure*}
    \centering
    \subfigure[]{
       \includegraphics[scale=0.35]{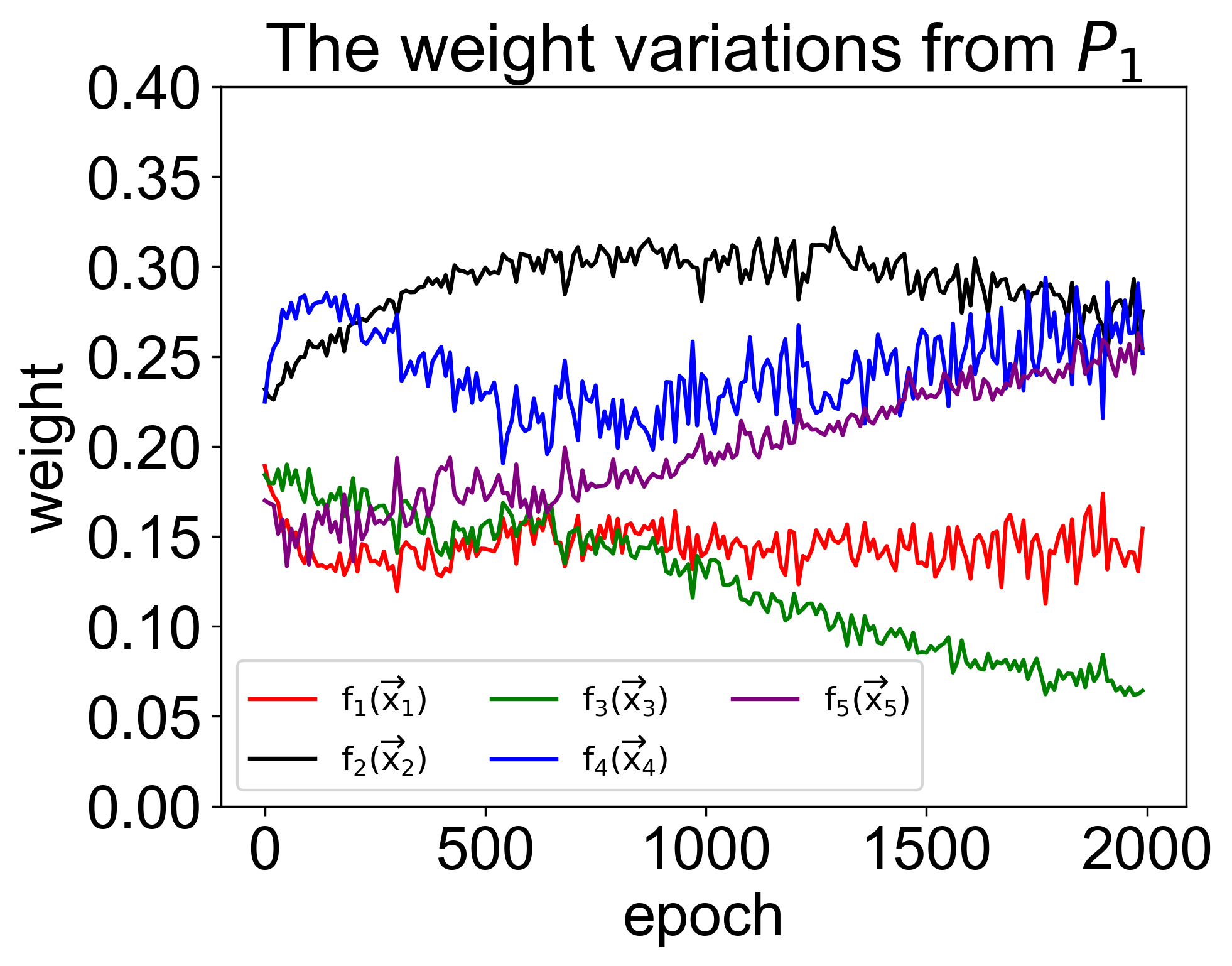}
    }
    \quad
    \subfigure[]{
       \includegraphics[scale=0.35]{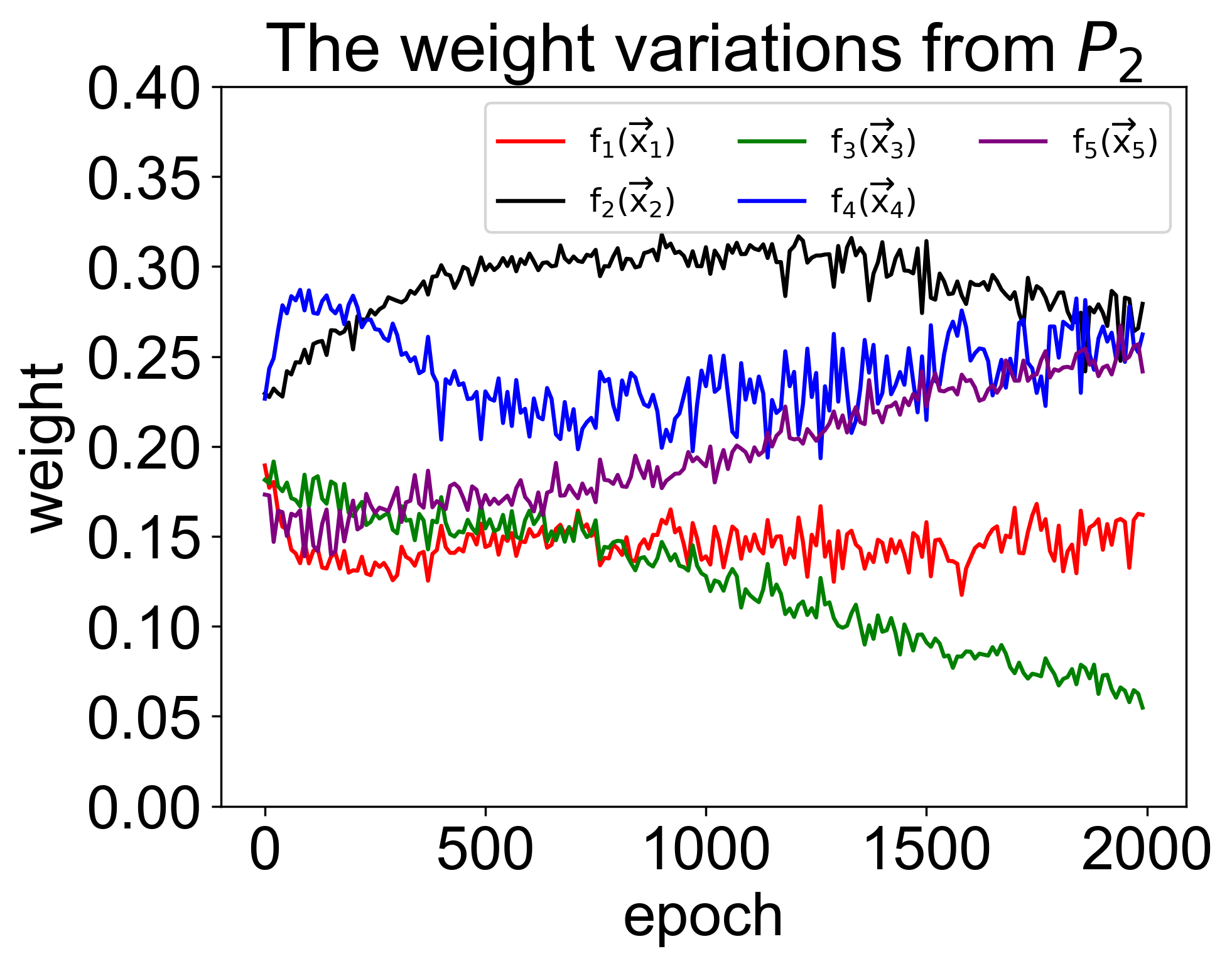}
    }\\
    \subfigure[]{
       \includegraphics[scale=0.35]{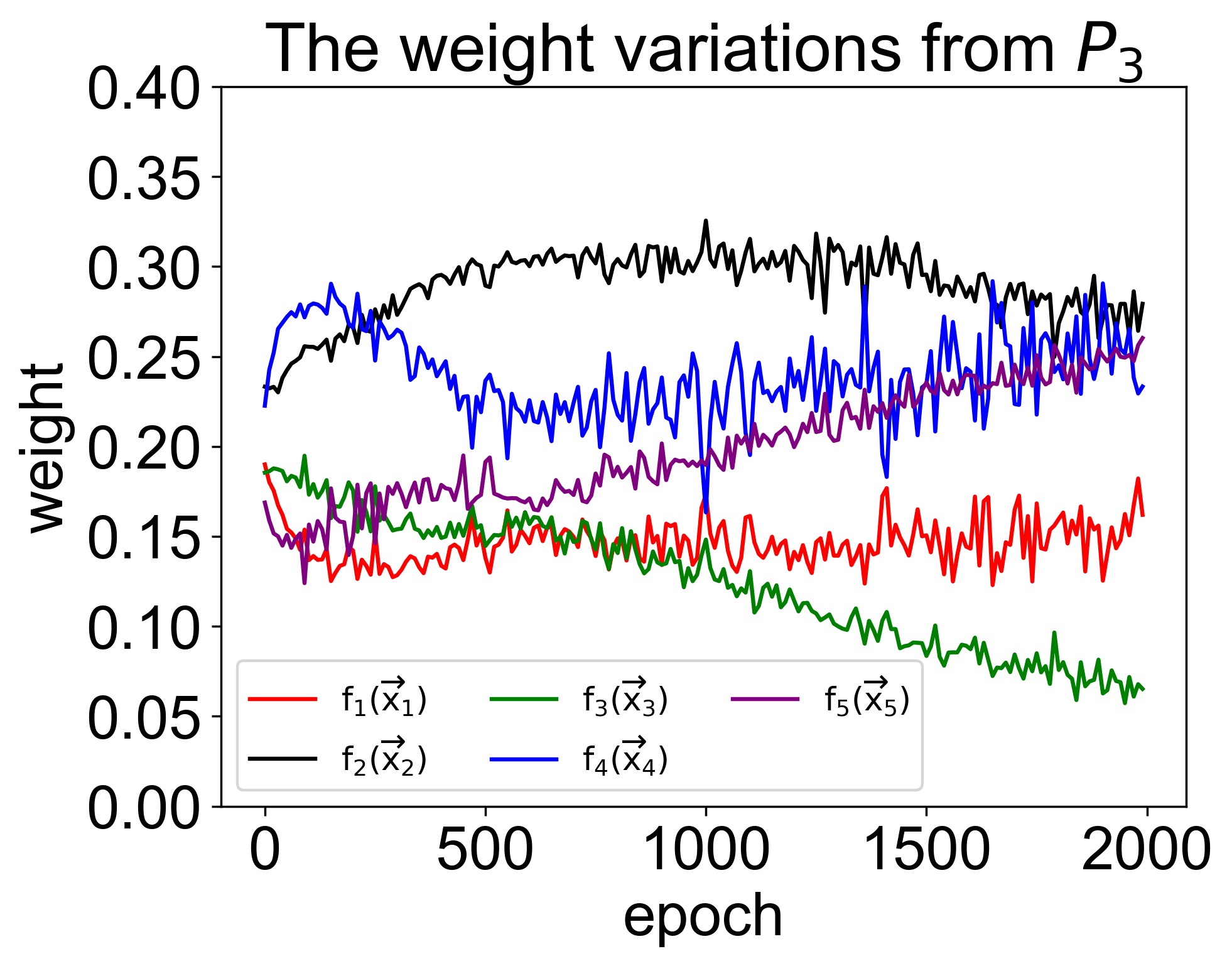}
    }\quad
    \subfigure[]{
       \includegraphics[scale=0.35]{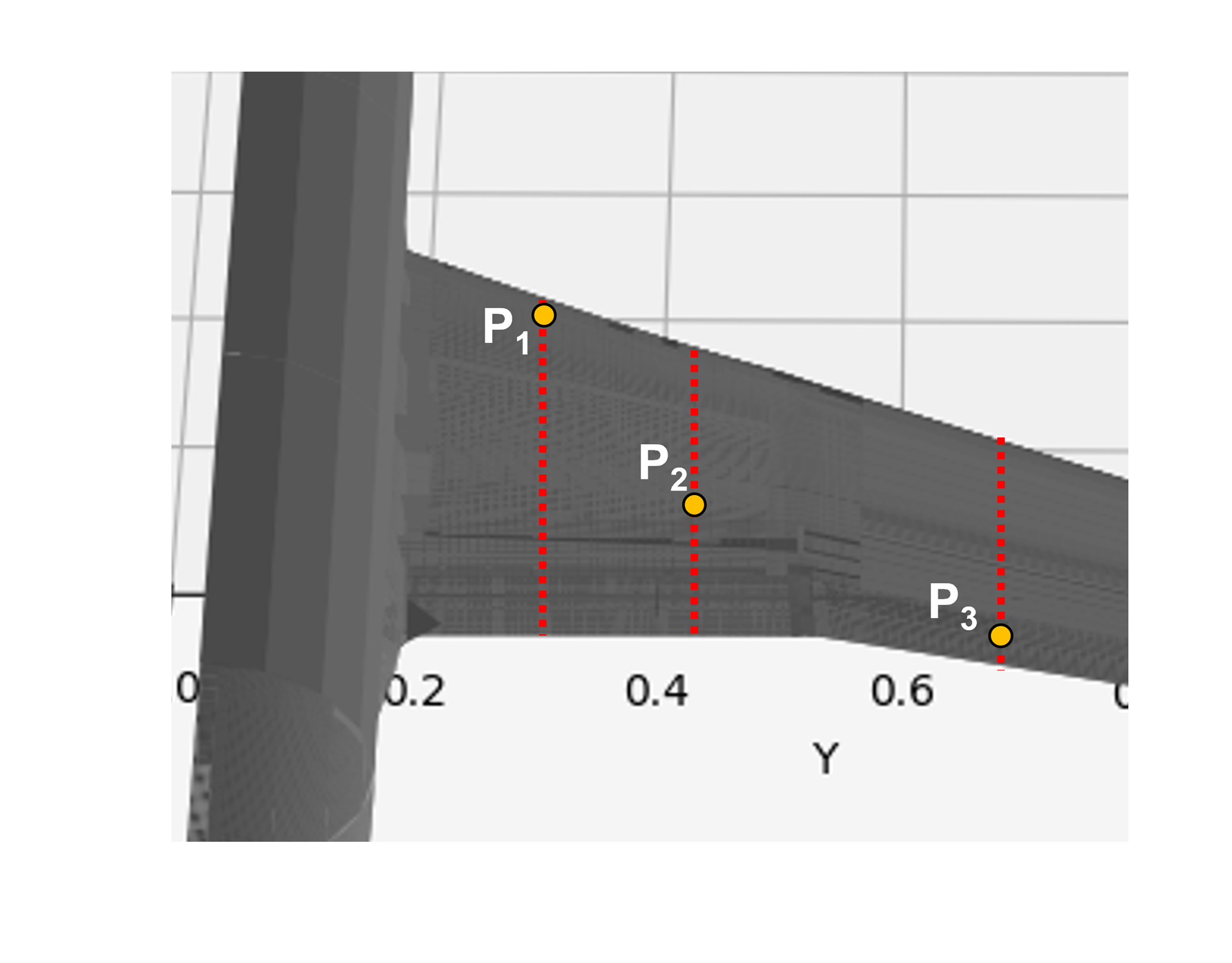}
    }\quad
    \caption{The weights of neural representations for three random points with $AoA=18.5$°.}
    \label{fig_weights_18.5}
\end{figure*}

\begin{figure*}
    \centering
    \subfigure[]{
       \includegraphics[scale=0.35]{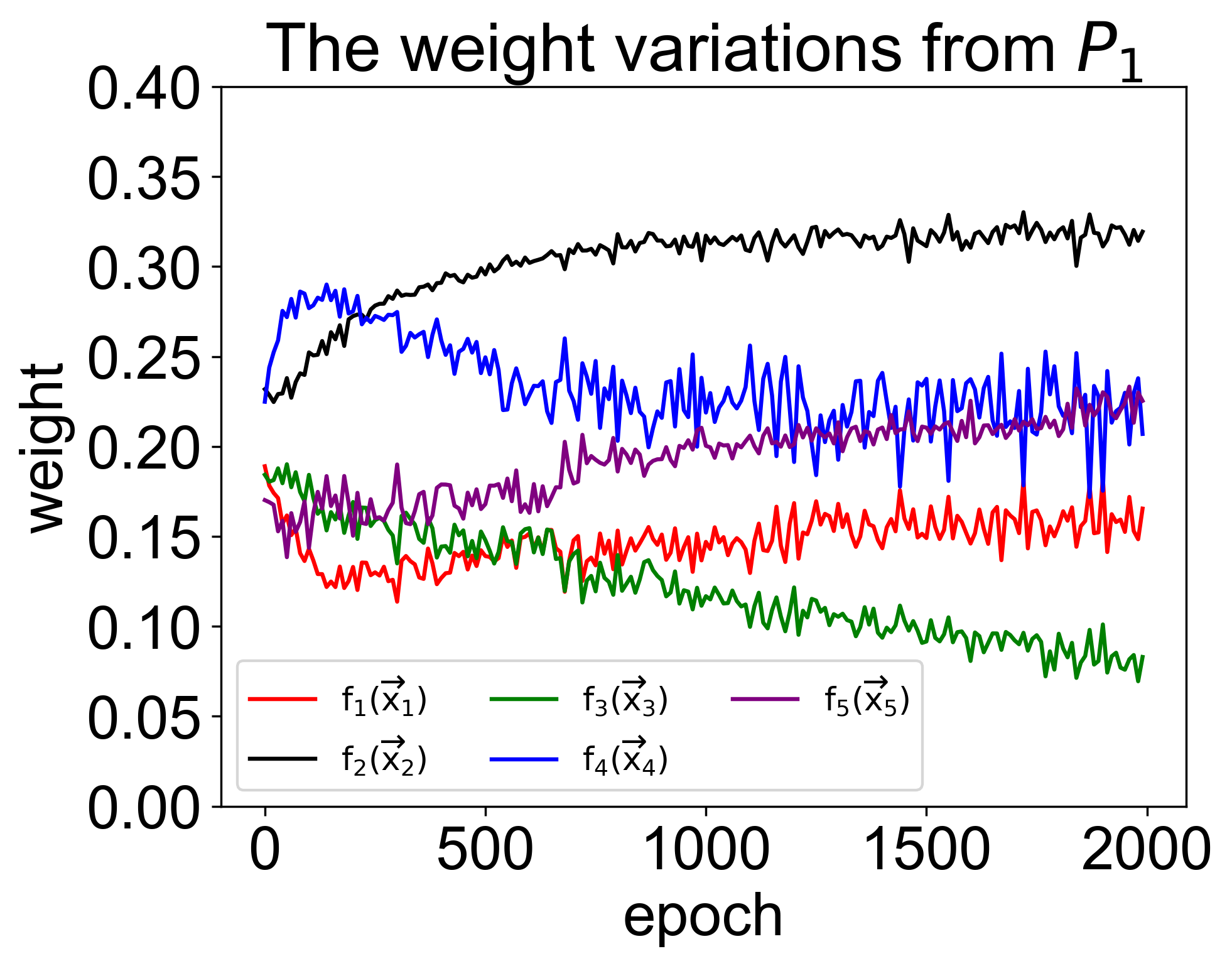}
    }
    \quad
    \subfigure[]{
       \includegraphics[scale=0.35]{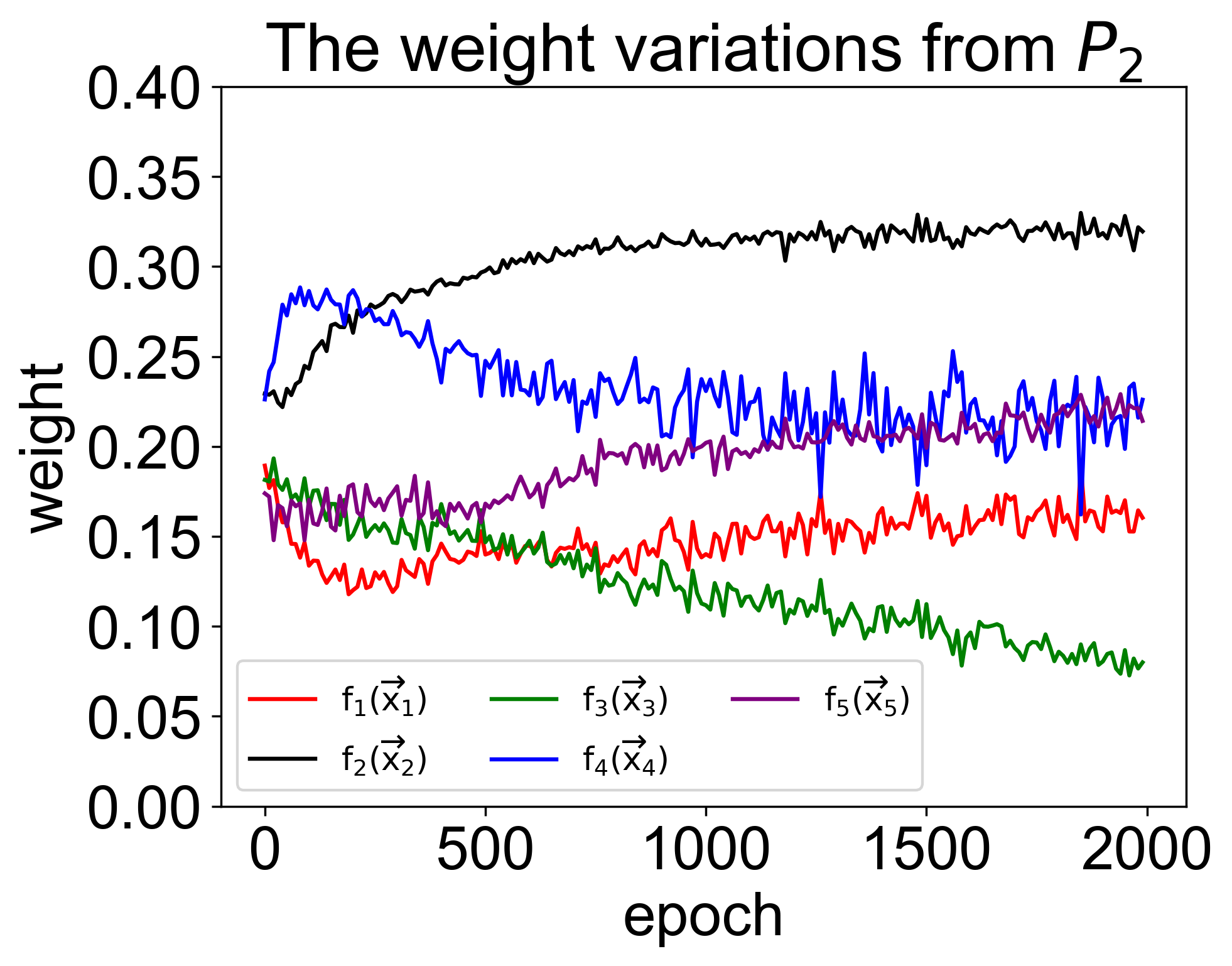}
    }\\
    \subfigure[]{
       \includegraphics[scale=0.35]{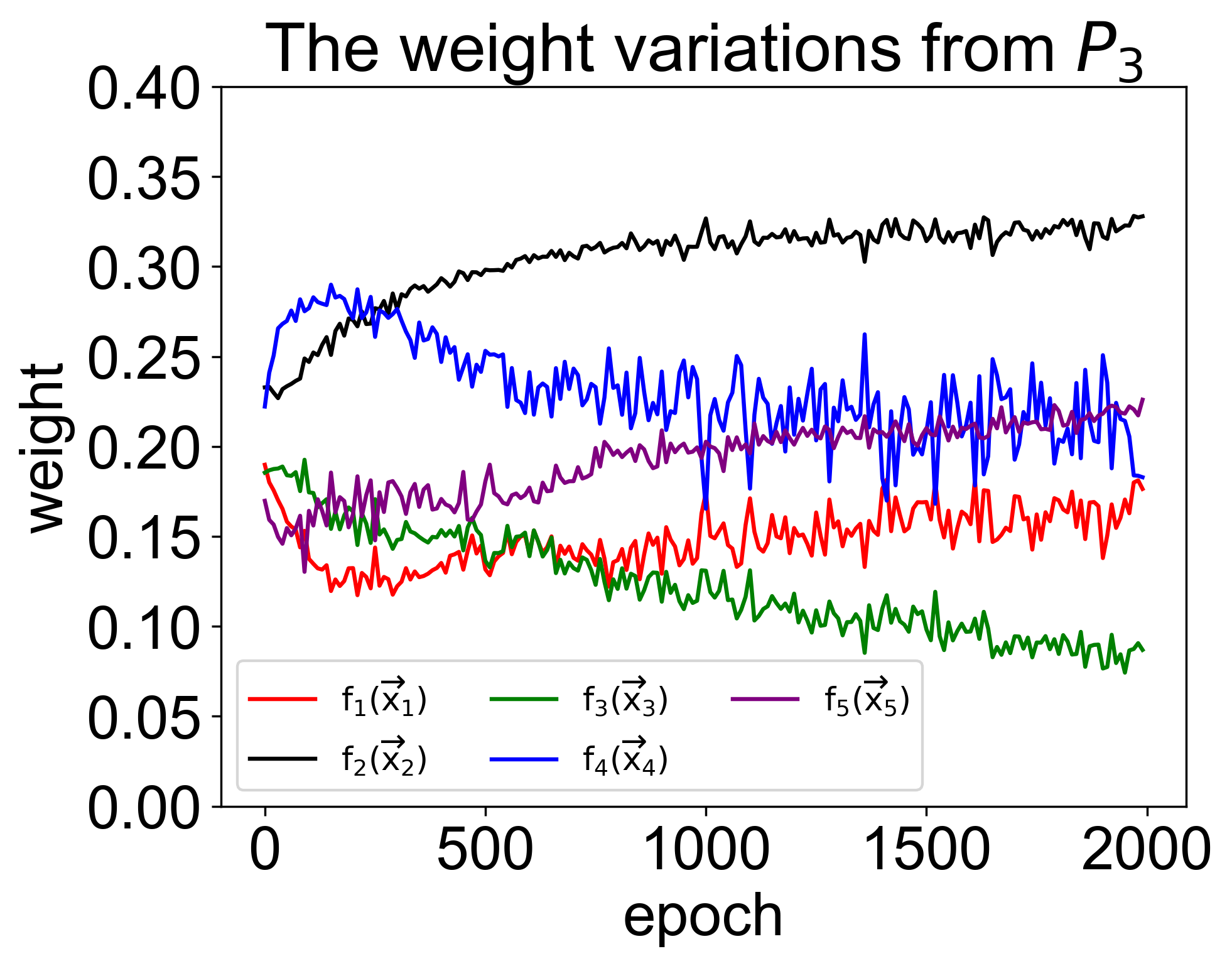}
    }\quad
    \caption{The weights of neural representations for three random points with $AoA=20$°.}
    \label{fig_weights_20}
\end{figure*}

\subsection{Analysis: The Contribution of Riemannian Geometric Features}
In this section, we analyze the importance of five neural representations during the training process of RGFiL\_2. Figure. \ref{fig_weights_18.5} shows the weight variations of neural representations (i.e., $f_1(\vec{\mathbf{x}}_1) \sim f_5(\vec{\mathbf{x}}_5)$) for three random points $P_1$, $P_2$ and $P_3$ on the DLR-F11 manifold with $AoA=18.5$°. Subgraphs (a) to (c) show the weight variations of neural representations on $P_1$, $P_2$ and $P_3$, respectively. Subgraph (d) shows the position of $P_1$, $P_2$ and $P_3$ on the DLR-F11 manifold. When epoch $\in [1000,2000]$, we see that 1) the weight of $f_2(\vec{\mathbf{x}}_2)$ decreases, while the weights of both $f_4(\vec{\mathbf{x}}_4)$ and $f_5(\vec{\mathbf{x}}_5)$ increase; 2) the weight of $f_1(\vec{\mathbf{x}}_1)$ does not significantly change; 3) the weights of $f_3(\vec{\mathbf{x}}_3)$ decreases.

Figure. \ref{fig_weights_20} shows the weight variations of $f_1(\vec{\mathbf{x}}_1) \sim f_5(\vec{\mathbf{x}}_5)$ for $P_1$, $P_2$ and $P_3$ on the DLR-F11 manifold with $AoA=20$°. We see that 1) the weight of $f_2(\vec{\mathbf{x}}_2)$ is biggest; 2) the weights of $f_4(\vec{\mathbf{x}}_4)$ and $f_5(\vec{\mathbf{x}}_5)$ are always bigger than those of $f_1(\vec{\mathbf{x}}_1)$ and $f_3(\vec{\mathbf{x}}_3)$; 3) the weights of $f_1(\vec{\mathbf{x}}_1)$, $f_2(\vec{\mathbf{x}}_2)$, $f_4(\vec{\mathbf{x}}_4)$ and $f_5(\vec{\mathbf{x}}_5)$ tend to be stable as the number of iterations increases, but the weight of $f_3(\vec{\mathbf{x}}_3)$ decreases as the number of iterations increases.

Theoretically, the Riemannian metric $g_{ij}$ is an inner product on the tangent space of a manifold. The $g_{ij}$ forms the basis for other Riemannian geometric features, for examples, the distance, angle, connection, and curvature \cite{xiang2023manifold}. The connection coefficient $\Gamma_{ij}^k$ is instrumental in defining covariant derivatives and parallel transport \cite{xie2013parallel}, establishing relationships between tangent spaces at different neighboring points on the aircraft manifold. Connections between different tangent spaces play a crucial role in representing the intrinsic geometric integrity of the aircraft manifold. Scalar curvature $S$ is employed to reflect the bending and deformation properties of the wing surface \cite{wang2023mixed}. The Riemannian geometric features are used to represent the intrinsic geometric properties of the 3D DLR-F11 aircraft, whcih explains why the RGFiL shows better accuracy than other methods. Notably, the geometric meanings of $\Gamma_{ij}^k$ and $S$ are more clear and suitable for aircrafts than that of $g_{ij}$, thus the RGFiL keeps reducing the weight of $f_3(\vec{\mathbf{x}}_3)$ during the training process.

\section{Conclusion and Future Work}
\label{section_conclusion}
In this paper, we propose RGFiL that incorporating Riemannian geometric features from an arbitrary point and its 8 neighbors to predict the $C_P$ distributions on a 3D aircraft. We demonstrate that the geometry of a 3D aircraft can be described by a piecewise smooth manifold, and further demonstrate that three Riemannian geometric features (Riemannian metric, connection and curvature) can be used to represent the geometric integrity of the 3D aircraft. Experimental results show that RGFiL reduces the average prediction MSE of $C_{P}$ by 15.00\% for the DLR-F11 aircraft test set.

In the future, we will build stronger relations between Riemannian geometry features and deep learning, for example the connection can be used to control the parallel transport of velocity vectors on aircraft manifold, which can be used to explore the internal mechanism of flow field variations.

\bibliographystyle{IEEEtran}
\bibliography{main}

\begin{thebibliography}{10}
\providecommand{\url}[1]{#1}
\csname url@samestyle\endcsname
\providecommand{\newblock}{\relax}
\providecommand{\bibinfo}[2]{#2}
\providecommand{\BIBentrySTDinterwordspacing}{\spaceskip=0pt\relax}
\providecommand{\BIBentryALTinterwordstretchfactor}{4}
\providecommand{\BIBentryALTinterwordspacing}{\spaceskip=\fontdimen2\font plus
\BIBentryALTinterwordstretchfactor\fontdimen3\font minus
  \fontdimen4\font\relax}
\providecommand{\BIBforeignlanguage}[2]{{%
\expandafter\ifx\csname l@#1\endcsname\relax
\typeout{** WARNING: IEEEtran.bst: No hyphenation pattern has been}%
\typeout{** loaded for the language `#1'. Using the pattern for}%
\typeout{** the default language instead.}%
\else
\language=\csname l@#1\endcsname
\fi
#2}}
\providecommand{\BIBdecl}{\relax}
\BIBdecl

\bibitem{yang2023inverse}
S.~Yang, S.~Lee, and K.~Yee, ``Inverse design optimization framework via a
  two-step deep learning approach: application to a wind turbine airfoil,''
  \emph{Engineering with Computers}, vol.~39, no.~3, pp. 2239--2255, 2023.

\bibitem{zahn2023prediction}
R.~Zahn, A.~Weiner, and C.~Breitsamter, ``Prediction of wing buffet pressure
  loads using a convolutional and recurrent neural network framework,''
  \emph{CEAS Aeronautical Journal}, pp. 1--17, 2023.

\bibitem{zuo2023fast}
K.~Zuo, Z.~Ye, W.~Zhang, X.~Yuan, and L.~Zhu, ``Fast aerodynamics prediction of
  laminar airfoils based on deep attention network,'' \emph{Physics of Fluids},
  vol.~35, no.~3, 2023.

\bibitem{zhang2024leave}
Y.~Zhang, Y.~Li, X.~Liu, W.~Deng \emph{et~al.}, ``Leave no stone unturned: Mine
  extra knowledge for imbalanced facial expression recognition,''
  \emph{Advances in Neural Information Processing Systems}, vol.~36, 2024.

\bibitem{yu2022point}
X.~Yu, L.~Tang, Y.~Rao, T.~Huang, J.~Zhou, and J.~Lu, ``Point-bert:
  Pre-training 3d point cloud transformers with masked point modeling,'' in
  \emph{Proceedings of the IEEE/CVF conference on computer vision and pattern
  recognition}, 2022, pp. 19\,313--19\,322.

\bibitem{baker2024learning}
P.~Baker, ``Learning word representations with projective geometry,'' Ph.D.
  dissertation, Universit{\'e} d'Ottawa/University of Ottawa, 2024.

\bibitem{alwadee2024latup}
E.~J. Alwadee, X.~Sun, Y.~Qin, and F.~C. Langbein, ``Latup-net: A lightweight
  3d attention u-net with parallel convolutions for brain tumor segmentation,''
  \emph{arXiv preprint arXiv:2404.05911}, 2024.

\bibitem{liu2024simultaneous}
H.~Liu, D.~Wei, D.~Lu, X.~Tang, L.~Wang, and Y.~Zheng, ``Simultaneous alignment
  and surface regression using hybrid 2d--3d networks for 3d coherent layer
  segmentation of retinal oct images with full and sparse annotations,''
  \emph{Medical Image Analysis}, vol.~91, p. 103019, 2024.

\bibitem{zhang2024data}
S.~Zhang and W.~Jiang, ``Data-informed geometric space selection,''
  \emph{Advances in Neural Information Processing Systems}, vol.~36, 2024.

\bibitem{mettes2024hyperbolic}
P.~Mettes, M.~Ghadimi~Atigh, M.~Keller-Ressel, J.~Gu, and S.~Yeung,
  ``Hyperbolic deep learning in computer vision: A survey,''
  \emph{International Journal of Computer Vision}, pp. 1--25, 2024.

\bibitem{cao2024knowledge}
J.~Cao, J.~Fang, Z.~Meng, and S.~Liang, ``Knowledge graph embedding: A survey
  from the perspective of representation spaces,'' \emph{ACM Computing
  Surveys}, vol.~56, no.~6, pp. 1--42, 2024.

\bibitem{deng2023prediction}
Z.~Deng, J.~Wang, H.~Liu, H.~Xie, B.~Li, M.~Zhang, T.~Jia, Y.~Zhang, Z.~Wang,
  and B.~Dong, ``Prediction of transonic flow over supercritical airfoils using
  geometric-encoding and deep-learning strategies,'' \emph{arXiv preprint
  arXiv:2303.03695}, 2023.

\bibitem{wang2023airfoil}
Y.~Wang, K.~Shimada, and A.~Barati~Farimani, ``Airfoil gan: encoding and
  synthesizing airfoils for aerodynamic shape optimization,'' \emph{Journal of
  Computational Design and Engineering}, vol.~10, no.~4, pp. 1350--1362, 2023.

\bibitem{xiang2023manifold}
Y.~Xiang, L.~Hu, G.~Zhang, J.~Zhang, and W.~Wang, ``A manifold-based airfoil
  geometric-feature extraction and discrepant data fusion learning method,''
  \emph{IEEE Transactions on Aerospace and Electronic Systems}, 2023.

\bibitem{catalani2023comparative}
G.~Catalani, D.~Costero, M.~Bauerheim, L.~Zampieri, V.~Chapin, N.~Gourdain, and
  P.~Baqu{\'e}, ``A comparative study of learning techniques for the
  compressible aerodynamics over a transonic rae2822 airfoil,'' \emph{Computers
  \& Fluids}, vol. 251, p. 105759, 2023.

\bibitem{jacob2021deep}
S.~J. Jacob, M.~Mrosek, C.~Othmer, and H.~K{\"o}stler, ``Deep learning for
  real-time aerodynamic evaluations of arbitrary vehicle shapes,'' \emph{arXiv
  preprint arXiv:2108.05798}, 2021.

\bibitem{li2021deep}
J.~Li and M.~Zhang, ``On deep-learning-based geometric filtering in aerodynamic
  shape optimization,'' \emph{Aerospace Science and Technology}, vol. 112, p.
  106603, 2021.

\bibitem{selig2011uiuc}
M.~Selig, ``Uiuc airfoil data site, department of aerospace engineering,''
  \emph{Urbana, Illinois: University of Illinois,(Jan 2007) www. ae. uiuc.
  edu/m-selig/ads. html. RL Fearn,“Airfoil Aerodynamics Using Panel
  Methods,” The Mathematica Journal}, pp. 10--4, 2011.

\bibitem{han2022survey}
K.~Han, Y.~Wang, H.~Chen, X.~Chen, J.~Guo, Z.~Liu, Y.~Tang, A.~Xiao, C.~Xu,
  Y.~Xu \emph{et~al.}, ``A survey on vision transformer,'' \emph{IEEE
  transactions on pattern analysis and machine intelligence}, vol.~45, no.~1,
  pp. 87--110, 2022.

\bibitem{qu2022predicting}
X.~Qu, Z.~Liu, B.~Yu, W.~An, X.~Liu, and H.~Lyu, ``Predicting pressure
  coefficients of wing surface based on the transfer of spatial dependency,''
  \emph{AIP Advances}, vol.~12, no.~5, 2022.

\bibitem{lei2021deeplearning}
R.~Lei, J.~Bai, H.~Wang, B.~Zhou, and M.~Zhang, ``Deep learning based
  multistage method for inverse design of supercritical airfoil,''
  \emph{Aerospace Science and Technology}, vol. 119, p. 107101, 2021.

\bibitem{xiong2023point}
F.~Xiong, L.~Zhang, H.~Xiao, and R.~Chengkun, ``A point cloud deep neural
  network metamodel method for aerodynamic prediction,'' \emph{Chinese Journal
  of Aeronautics}, vol.~36, no.~4, pp. 92--103, 2023.

\bibitem{zhang2023airfoil}
B.~Zhang, ``Airfoil-based convolutional autoencoder and long short-term memory
  neural network for predicting coherent structures evolution around an
  airfoil,'' \emph{Computers \& Fluids}, vol. 258, p. 105883, 2023.

\bibitem{chen2022learning}
Q.~Chen, P.~Pope, and M.~Fuge, ``Learning airfoil manifolds with optimal
  transport,'' in \emph{AIAA SCITECH 2022 Forum}, 2022, p. 2352.

\bibitem{van2022effect}
A.~Van~Slooten and M.~Fuge, ``Effect of optimal geometries and performance
  parameters on airfoil latent space dimension,'' in \emph{International Design
  Engineering Technical Conferences and Computers and Information in
  Engineering Conference}, vol. 86229.\hskip 1em plus 0.5em minus 0.4em\relax
  American Society of Mechanical Engineers, 2022, p. V03AT03A005.

\bibitem{wang2021airfoil}
Y.~Wang, K.~Shimada, and A.~B. Farimani, ``Airfoil gan: Encoding and
  synthesizing airfoils foraerodynamic-aware shape optimization,'' \emph{arXiv
  preprint arXiv:2101.04757}, 2021.

\bibitem{hu2022aerodynamic}
L.~Hu, Y.~Xiang, J.~Zhang, Z.~Shi, and W.~Wang, ``Aerodynamic data predictions
  based on multi-task learning,'' \emph{Applied Soft Computing}, vol. 116, p.
  108369, 2022.

\bibitem{lee2012smooth}
J.~M. Lee and J.~M. Lee, \emph{Smooth manifolds}.\hskip 1em plus 0.5em minus
  0.4em\relax Springer, 2012.

\bibitem{RiemannianChen}
w.~Chen, ``Riemannian metric,'' in \emph{Introduction to Riemann geometry in
  Chinese}, 1st~ed., j.~Zhang, Ed.\hskip 1em plus 0.5em minus 0.4em\relax
  Beijing: Beijing Normal University Publishing House, 2004, pp. 83--90.

\bibitem{van2023stochastic}
G.~G. van~der Vleuten, F.~Toschi, W.~H. Schilders, and A.~Corbetta,
  ``Stochastic fluctuations of diluted pedestrian dynamics along curved
  paths,'' \emph{arXiv preprint arXiv:2306.12090}, 2023.

\bibitem{khatsymovsky2019discrete}
V.~Khatsymovsky, ``On the discrete christoffel symbols,'' \emph{International
  Journal of Modern Physics A}, vol.~34, no.~30, p. 1950186, 2019.

\bibitem{pennec2019riemannian}
X.~Pennec, S.~Sommer, and T.~Fletcher, \emph{Riemannian geometric statistics in
  medical image analysis}.\hskip 1em plus 0.5em minus 0.4em\relax Academic
  Press, 2019.

\bibitem{frost2023lie}
H.~Frost, C.~R. Mafra, and L.~Mason, ``A lie bracket for the momentum kernel,''
  \emph{Communications in Mathematical Physics}, vol. 402, no.~2, pp.
  1307--1343, 2023.

\bibitem{munteanu2023comparison}
O.~Munteanu and J.~Wang, ``Comparison theorems for 3d manifolds with scalar
  curvature bound,'' \emph{International Mathematics Research Notices}, vol.
  2023, no.~3, pp. 2215--2242, 2023.

\bibitem{Zhang2022amultitask}
J.~Zhang, G.~Zhang, Y.~Cheng, L.~Hu, Y.~Xiang, and W.~Wang, ``A multi-task
  learning method for large discrepant aerodynamic data,'' \emph{Acta
  Aerodynamica Sinica}, vol.~40, pp. 64--72, 2022.

\bibitem{xin2022surrogate}
D.~Xin, J.~Zeng, and K.~Xue, ``Surrogate drag model of non-spherical fragments
  based on artificial neural networks,'' \emph{Powder Technology}, vol. 404, p.
  117412, 2022.

\bibitem{coder2015overflow}
J.~G. Coder, ``Overflow analysis of the dlr-f11 high-lift configuration
  including transition modeling,'' \emph{Journal of Aircraft}, vol.~52, no.~4,
  pp. 1082--1097, 2015.

\bibitem{yamin2022neural}
M.~Yamin and Z.~A.~K. Ramadhan, ``Neural network approach for predicting
  aerodynamic performance of naca airfoil at low reynolds number,''
  \emph{Jurnal Polimesin}, vol.~20, no.~2, pp. 190--193, 2022.

\bibitem{ivan2023empirical}
I.~Vallés-Pérez, E.~Soria-Olivas, M.~Martínez-Sober, A.~J. Serrano-López,
  J.~Vila-Francés, and J.~Gómez-Sanchís, ``Empirical study of the modulus as
  activation function in computer vision applications,'' \emph{Engineering
  Applications of Artificial Intelligence}, vol. 120, p. 105863, 2023.

\bibitem{xie2013parallel}
Q.~Xie, S.~Kurtek, H.~Le, and A.~Srivastava, ``Parallel transport of
  deformations in shape space of elastic surfaces,'' in \emph{Proceedings of
  the IEEE International Conference on Computer Vision}, 2013, pp. 865--872.

\bibitem{wang2023mixed}
J.~Wang, Y.~Shi, H.~Yu, X.~Wang, Z.~Yan, and F.~Kong, ``Mixed-curvature
  manifolds interaction learning for knowledge graph-aware recommendation,'' in
  \emph{Proceedings of the 46th International ACM SIGIR Conference on Research
  and Development in Information Retrieval}, 2023, pp. 372--382.

\end{thebibliography}

\begin{IEEEbiography}[{\includegraphics[width=1in,height=1.25in,clip,keepaspectratio]{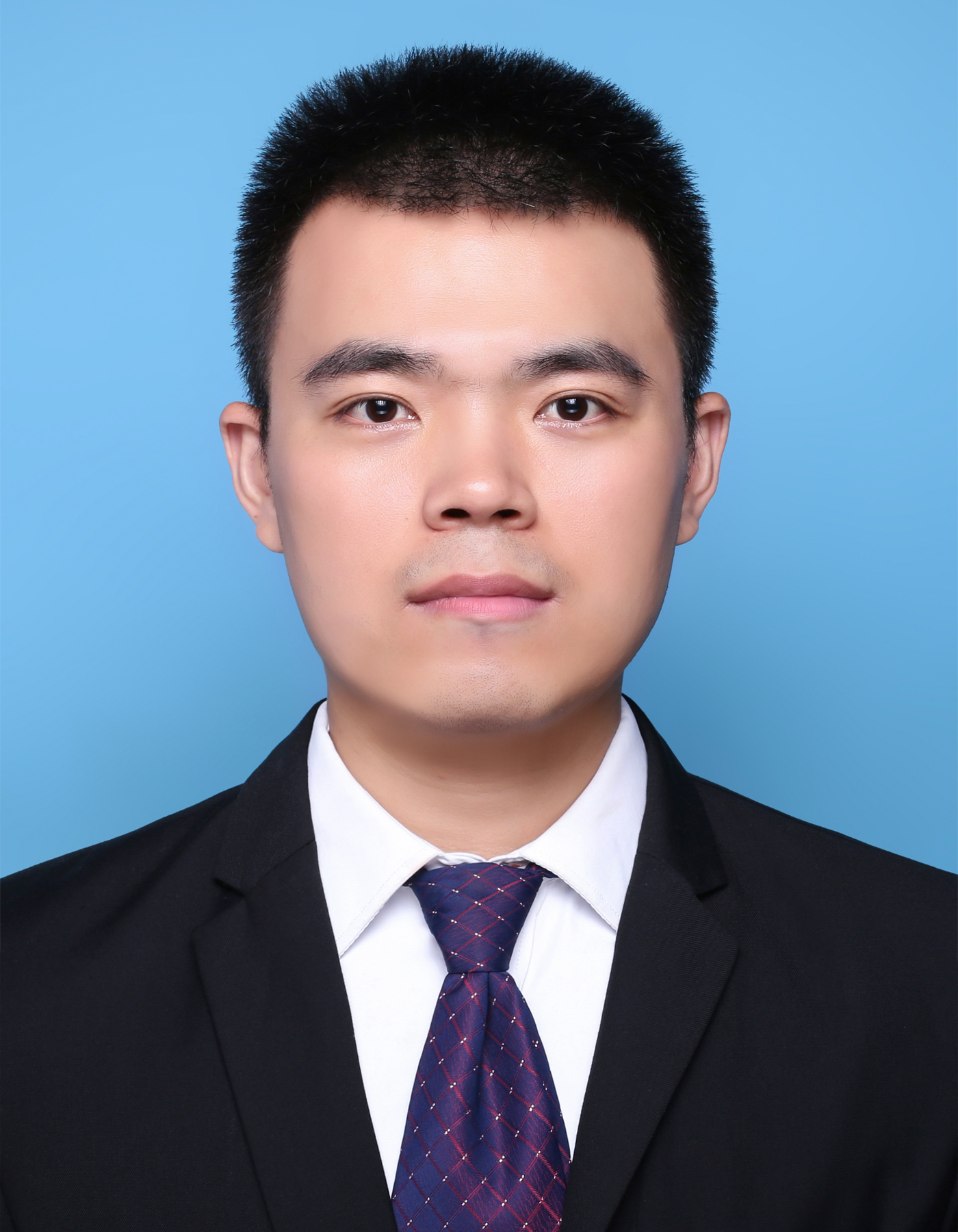}}]{Liwei Hu} received a B.S. degree in software engineering from the School of Software, Hebei Normal University, Shijiazhuang, Hebei, China, in 2014, and M.S. and Ph.D. degrees from the University of Electronic Science and Technology (UESTC), Chengdu, China, in 2018 and 2023, respectively. He has been a lecturer in School of Information Science and Engineering, Hebei University of Science and Technology since 2024. His research fields include aerodynamic data modeling, deep learning and pattern recognition.
\end{IEEEbiography}

\begin{IEEEbiography}[{\includegraphics[width=1in,height=1.25in,clip,keepaspectratio]{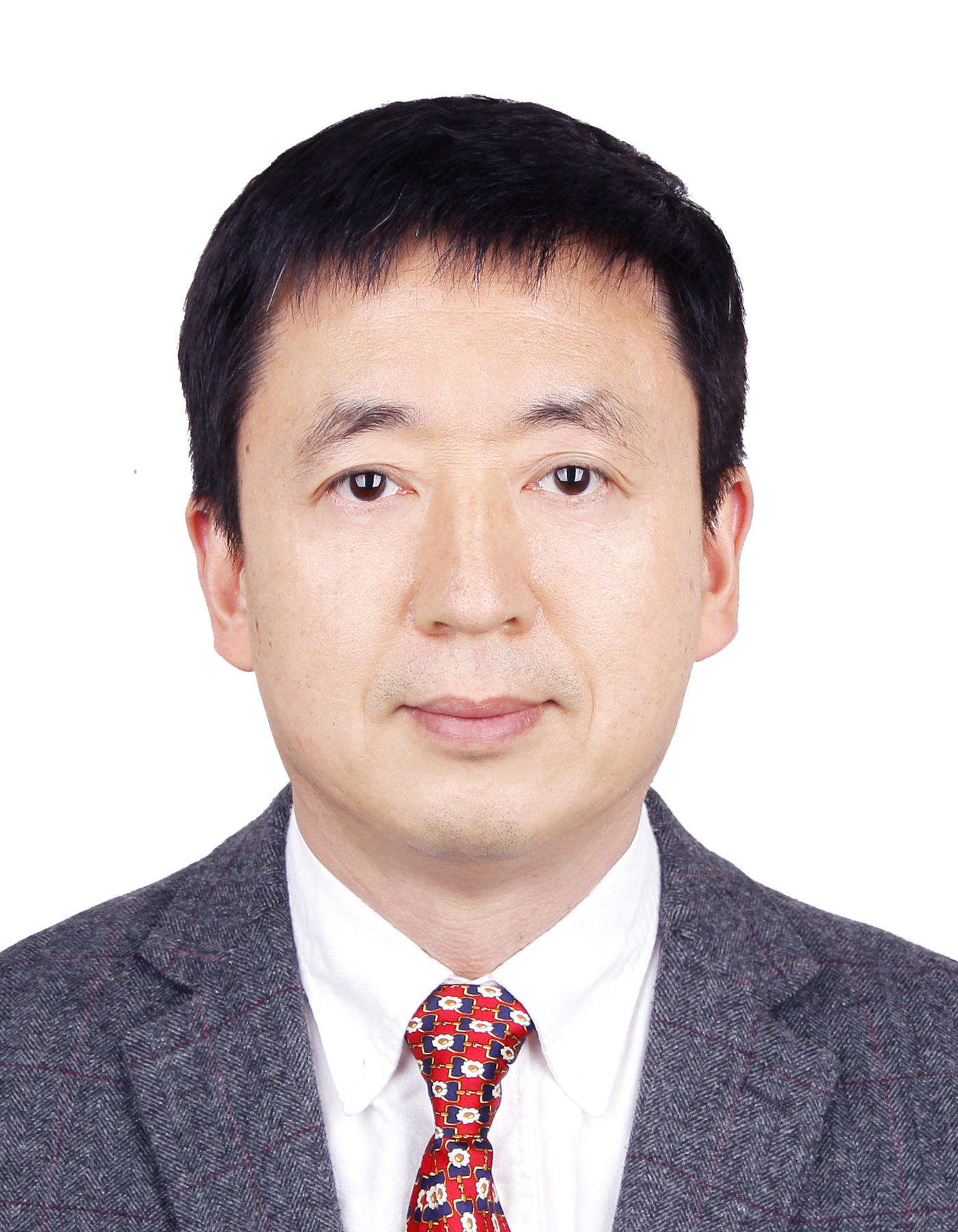}}]{Wenyong Wang} received a B.S. degree in computer science from BeiHang University, Beijing, China, in 1988 and M.S. and Ph.D. degrees from the University of Electronic Science and Technology (UESTC), Chengdu, China, in 1991 and 2011, respectively. He has been a professor in computer science and engineering at UESTC since 2006. He is also a special-term professor at Macau University of Science and Technology, a senior member of the Chinese Computer Federation, a member of the expert board of the China Education and Research Network (CERNET) and China next generation internet. His main research interests include next generation Internet, software-defined networks, and software engineering.
\end{IEEEbiography}

\begin{IEEEbiography}[{\includegraphics[width=1in,height=1.25in,clip,keepaspectratio]{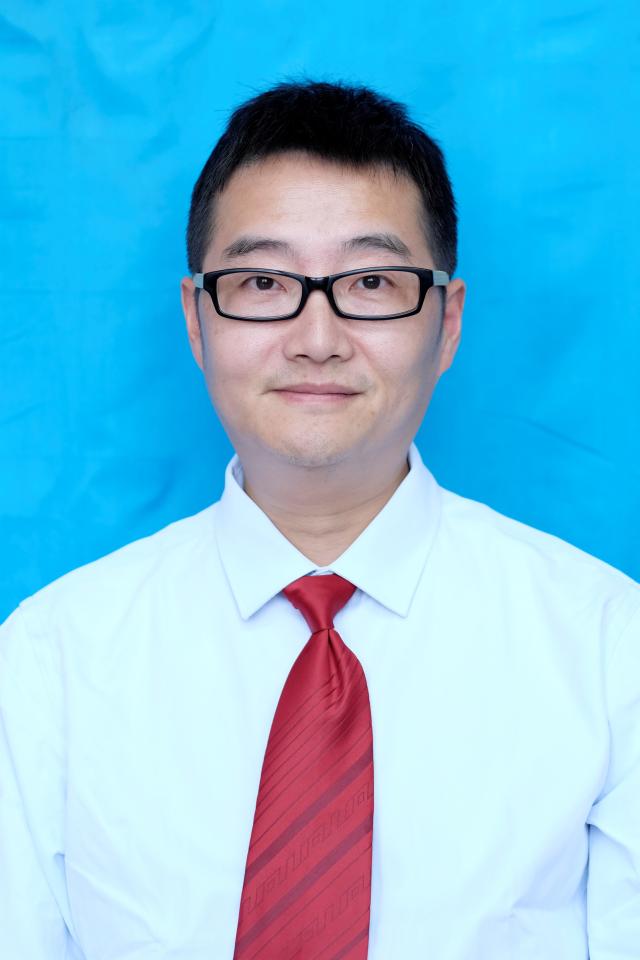}}]{Yu Xiang} received B.S, M.S. and Ph.D. degrees from the University of Electronic Science and Technology of China (UESTC), Chengdu, Sichuan, China, in 1995, 1998 and 2003, respectively. He joined the UESTC in 2003 and became an associate professor in 2006. From 2014-2015, he was a visiting scholar at the University of Melbourne, Australia. His current research interests include computer networks, intelligent transportation systems and deep learning.
\end{IEEEbiography}

\begin{IEEEbiography}[{\includegraphics[width=1in,height=1.25in,clip,keepaspectratio]{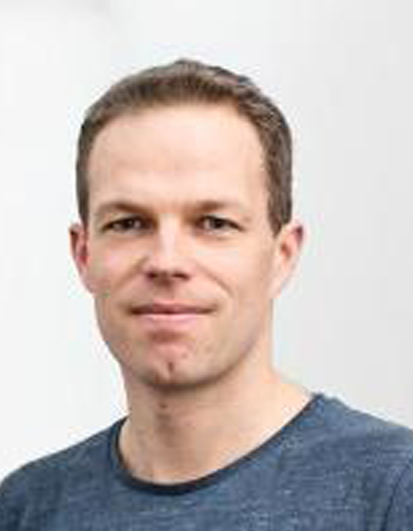}}]{Stefan Sommer} received his M.S. in mathematics in 2008, and his Ph.D. in computer science in 2012 from the University of Copenhagen. He is currently Professor at the Department of Computer Science, University of Copenhagen. His research interests include shape analysis theory and applications in science and engineering, and statistics and machine learning on data with complex, geometric structure (geometric statistics).
\end{IEEEbiography}

\end{document}